\documentclass[journal]{ieeetran}  
\IEEEoverridecommandlockouts                              

\usepackage{graphics} 
\usepackage{amsmath} 
\usepackage{amssymb}  
\usepackage{siunitx}
\usepackage{mathtools}
\usepackage{booktabs}
\usepackage[ruled]{algorithm2e}
\usepackage{subcaption} 
\usepackage[bottom=57pt,top=54pt, left=54pt, right=54pt]{geometry}   

\usepackage{adjustbox}
\usepackage{pbox}
\usepackage{dblfloatfix}
\usepackage{graphicx}
\usepackage[dvipsnames]{xcolor}
\usepackage{hyperref}
\usepackage{caption}
\usepackage{subcaption}
\usepackage[nospace]{cite}

\usepackage{array,multirow,graphicx}
\usepackage{makecell}
\usepackage[dvipsnames]{xcolor}

\newcommand{\etal}{\textit{et~al.}}
\newcommand{\minisec}[1]{\textbf{#1}}
\definecolor{class_floor}{RGB}{43, 160, 4}
\definecolor{class_wall}{RGB}{158, 216, 229}
\definecolor{class_furniture}{RGB}{255, 127, 12}

\DeclareMathOperator{\exectime}{time}
\DeclareMathOperator{\subtree}{subtree}
\DeclareMathOperator{\treepath}{path}
\DeclareMathOperator{\visible}{visible}

\begin{document}

\title{
SC-Explorer: Incremental 3D Scene Completion for Safe and Efficient Exploration Mapping and Planning
}

\author{Lukas Schmid$^{1}$, Mansoor Nasir Cheema$^{2}$, Victor Reijgwart$^{1}$, Roland Siegwart$^1$, Federico Tombari$^{2,3}$, \\and Cesar Cadena$^{1}$%
\thanks{This work was supported by funding from the Microsoft Swiss Joint Research Center.}%
\thanks{$^1$ Autonomous Systems Lab, ETH Z\"urich, Z\"urich, Switzerland}%
\thanks{$^2$ Technical University of Munich, Munich, Germany}%
\thanks{$^3$ Google, Z\"urich, Switzerland}%
\thanks{{\tt\small schmluk@ethz.ch}}%
}%

\maketitle

\begin{abstract}
Exploration of unknown environments is a fundamental problem in robotics and an essential component in numerous applications of autonomous systems.
A major challenge in exploring unknown environments is that the robot has to plan with the limited information available at each time step.
While most current approaches rely on heuristics and assumption to plan paths based on these partial observations, we instead propose a novel way to integrate deep learning into exploration by leveraging 3D scene completion for informed, safe, and interpretable exploration mapping and planning.
Our approach, \emph{SC-Explorer}, combines scene completion using a novel incremental fusion mechanism and a newly proposed hierarchical multi-layer mapping approach, to guarantee safety and efficiency of the robot. 
We further present an informative path planning method, leveraging the capabilities of our mapping approach and a novel scene-completion-aware information gain.
While our method is generally applicable, we evaluate it in the use case of a Micro Aerial Vehicle (MAV).
We thoroughly study each component in high-fidelity simulation experiments using only mobile hardware, and show that our method can speed up coverage of an environment by 73\% compared to the baselines with only minimal reduction in map accuracy.
Even if scene completions are not included in the final map, we show that they can be used to guide the robot to choose more informative paths, speeding up the measurement of the scene with the robot's sensors by 35\%.
We validate our system on a fully autonomous MAV, showing rapid and reliable scene coverage even in a complex cluttered environment.
We make our methods available as open-source.

\end{abstract}


\begin{figure}
\centering
\includegraphics[width=\linewidth]{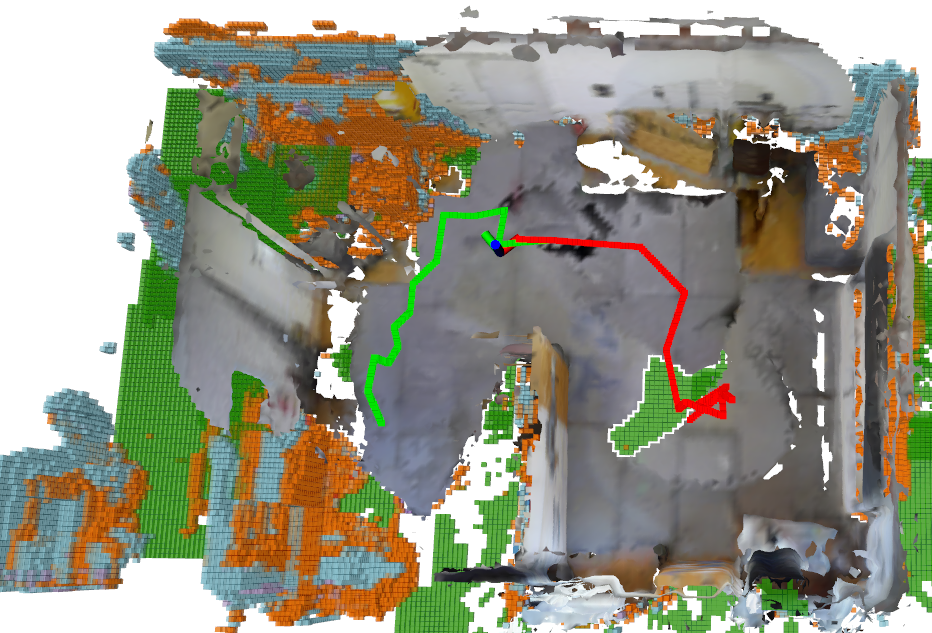}
\caption{Qualitative scene-completion-ware exploration results on a fully autonomous MAV. The mesh represents the measured 3D reconstruction, whereas voxels are non-measured areas that are predicted to be occupied by our scene completion network. The voxels are colored by class, indicating \textcolor{class_floor}{\textbf{floor}}, \textcolor{class_wall}{\textbf{wall}}, and \textcolor{class_furniture}{\textbf{furniture}}. 
The current pose of the robot is indicated as small axes, between the \textcolor{red}{\textbf{executed}} and \textcolor{green}{\textbf{currently planned}} path.
SC-Explorer is able to meaningfully fill in the holes in the reconstruction (e.g. below the start pose, top right corner), and predict the unknown areas (left) to which an intuitive exploration path is planned, even in this challenging out-of-training-distribution environment.}
\label{fig:kiwi_qualitative}
\end{figure}

\section{Introduction}
\label{sec:introduction}

The ability to autonomously explore and map an unknown environment is a crucial capability in robotics. 
It is essential for robot autonomy, and a prerequisite for a variety of applications, ranging from consumer robotics to autonomous surveying and inspection \cite{yoder2016autonomous,Nikolic2013AUS,Omari2014VisualII}, crop monitoring \cite{Khanna2015BeyondPC, popovic2020informative}, as well as search and rescue \cite{nevatia2008augmented, Colas20133DPP}.
In particular, Micro Aerial Vehicles (MAVs) have experienced widespread adoption of these tasks. 
Thus, while our method can be applied to any robot, without loss of generality, in this work we evaluate our approach in the use case of an aerial robot.

In volumetric exploration, the goal is to create a complete map of a previously unknown environment, and to do so quickly and safely. 
This is a challenging problem, as the robot needs to make a decision on where to move next based on the limited information available at every time step. 
Furthermore, the reconstructed maps are susceptible to occlusions, requiring the robot to thoroughly inspect every piece of the scene to observe all initially occluded areas.

To tackle the volumetric exploration problem, a broad variety of approaches have been proposed.
Most prominently, sampling-based planners \cite{Bircher2016RecedingH, Schmid20ActivePlanning ,dang2019graph} sample random candidate view poses and use sensor forward simulation, such as ray-casting, to determine the information gain and final utility of each candidate, based upon which the Next Best View (NBV) is chosen.
Frontier-based methods \cite{yamauchi1997frontier, cieslewski2017rapid} explicitly compute the boundary between observed and unknown space, termed the \emph{frontier}, from the map and use this to govern the robot motion.
While the majority of exploration approaches can be separated into sampling-based and frontier-based methods, also optimization-based \cite{shade_choosing_2011} or physics inspired methods \cite{shen_autonomous_2012} have been proposed.

More recently, advances in Deep learning (DL) methods have made a large impact on many domains of robotics \cite{sunderhauf2018limits}. 
However, their adaption to exploration planning has received less attention. 
Typically, active planning problems are addressed via Reinforcement Learning (RL) \cite{henderson2018deep, zhu2018, niroui2019, chen2019self}, which directly learns a mapping of the current state to an action. 
However, RL methods are known to be hard to train \cite{henderson2018deep} and thus typically require significant simplifications.
A different family of approaches use DL to improve the computation cost of exploration planning. 
These include imitation learning \cite{andersson2017deep, reinhart2020, Bai2017}, where a network is trained to reconstruct the actions of a teacher method.
Other approaches include the combination of classical sampling-based methods with learning, trying to learn the information gain \cite{hepp_learnscore_2018} or also the underlying informed distribution of viewpoints \cite{schmid2022fast}. 

However, all of these methods still suffer from the challenges of limited information and occlusions in exploration planning. 
Most importantly, the majority of the relevant information to choose the next action in exploration is within close proximity of the robot, which is mostly unknown space \cite{schmid2021glocal}. 
To overcome these limitations, we propose a novel way of combining deep learning with robotic exploration. 
In particular, we propose to leverage deep 3D Scene Completion (SC) \cite{palnet, roldao20223d} to predict the unknown areas of the partial observations based on structural and semantic priors, and investigate its potential impact on both mapping and planning.

We thus propose \emph{SC-Explorer}, a complete SC-aware mapping and planning pipeline.
Our system incrementally completes the surroundings of the robot to fill in unknown parts. 
We investigate different ways of fusing this information and propose a novel fusion mechanism tailored to SC.
The SC is then combined with the sensor measurements in a novel multi-layer mapping architecture to guarantee safety and efficiency of the robot even in the presence of erroneous predictions.
We further propose and investigate several information gain formulations to leverage this novel information in the map and drive exploration.
We experimentally study each component and find that SC can significantly improve mapping and planning performance via multiple mechanisms, but can also hinder exploration as erroneous predictions can block the robot and thus requires careful consideration as in our proposed approach.
We further experimentally study map accuracy, safety, and exploration speed trade-offs. 

In summary, we make the following contributions:
\begin{itemize}
    \item We propose a novel approach to integrate deep learning into exploration, leveraging incremental 3D scene completion for safe and interpretable mapping and planning.
    \item We present a novel method for SC-based mapping, consisting of a multi-layer map representation and a novel incremental SC prediction fusion method, enabling safe and efficient planning.
    \item We present a novel informative path planning system leveraging the capabilities of our multi-layer SC map and an SC-aware information gain for fast autonomous exploration planning.
\end{itemize}

We show in thorough experimental evaluation using high-fidelity simulations that our contributions can speed up coverage of the environment by 73\% compared to a classical and an SC-aware baseline, and even when SC are not included in the final map, they can be used to guide the robot speeding up the measurement of the environment with the sensor by 35\%. 
We further validate our approach on a physical robot, demonstrating the applicability and reliable performance of SC-Explorer on a fully autonomous MAV.
We make our methods available as open-source for the benefit of the community\footnote{Released at \url{https://github.com/ethz-asl/ssc_exploration} upon publication.}.

\section{Related Work} 
\label{sec:rel_work}

In this section, we first independently review the state of the art in exploration in Sec.~\ref{sec:rel_exploration} and 3D scene completion in Sec.~\ref{sec:rel_sc}, then discuss applications of learning in robotic exploration in Sec.~\ref{sec:rel_learning}.

\subsection{Classical Exploration Planning}
\label{sec:rel_exploration}

The majority of exploration approaches can be divided into sampling-based and frontier-based methods \cite{julia2012comparison}. 
These approaches have complementary properties and found success in different aspects of exploration planning.

In frontier-based planning, the boundaries of known space are explicitly computed from the map and used as navigation goal \cite{yamauchi1997frontier}. 
This has the advantage that all such frontiers are detected, guaranteeing global coverage, but can be computationally expensive in large volumetric maps \cite{Keidar2014frontier}. 
It is therefore rarely used in stand-alone fashion. 
A notable exception is Cieslewski \etal\cite{cieslewski2017rapid}, who proposes a frontier following method minimizing MAV steering and therefore increasing flight speed.

In sampling-based exploration, viewpoints are sampled in the configuration or map space of the robot, an information gain, cost, and combined final utility is computed for each viewpoint, and the best candidate is chosen for execution. 
This has the advantage that feasible paths can easily be sampled and volumetric gains of various kinds \cite{nbvp_uncertainty, nbvp_object_search, isler2016information, Schmid20ActivePlanning} can be computed via ray-casting. 
Due to the high efficiency of local paths, this has emerged as the dominant approach to exploration planning.

A common structure to organize viewpoints is in a Rapidly-exploring Random Tree (RRT) as introduced by Bircher \etal~\cite{Bircher2016RecedingH}, where all new viewpoints are added to the RRT, and only the first element of the best branch is executed.
This approach has found success in a variety of applications and local planners \cite{Bircher2016RecedingH, nbvp_uncertainty, nbvp_object_search, Selin_nbv_fron, history_nbvp}. The concept has been further extended to more general graphs by Dang \etal~\cite{dang2019graph, dang2020graph} and Schmid \etal~\cite{Schmid20ActivePlanning, schmid2021glocal} to allow for more flexibility through rewiring. 
While these methods can have considerable coverage, they focus primarily on local planning and complete coverage can not be guaranteed as the number of samples is finite.

To tackle this issue, a trend to split exploration into local and global planning problems has emerged, where typically a sampling-based method is employed for local planning.
Similar to before, global planning methods can be divided into two categories. 
First, the 'Library of Views' approach explicitly stores past view points that still have some information gain as candidates for global relocation, once the robot is stuck in a dead end \cite{history_nbvp, dang2019graph, gmm_multirobot_expl}. 
A slight modification is presented by Selin \etal~\cite{Selin_nbv_fron}, who uses a Gaussian Process (GP) to estimate this distribution of views.
Second, works combine sampling-based local planning with frontier-based global planning to guarantee coverage \cite{schmid2021glocal, respall2021fast}.

Lastly, a different way to combine both modalities is to use frontiers to guide the sampling process \cite{dai2020fast, kompis2021informed, meng_2stage_expl, zhou2021fuel}. 
These either use frontiers to generate informed viewpoint samples for classical sampling-based planning \cite{dai2020fast, kompis2021informed} or to compute an optimal traversal sequence by solving a Traveling Salesman Problem (TSP) \cite{meng_2stage_expl, zhou2021fuel}.

However, all of these methods plan only on the limited information available at each time step, primarily employing heuristics to estimate which parts of the environment can be seen from each view, and plan cumbersome paths to observe small holes resulting from occlusions.


\subsection{3D Scene Completion}
\label{sec:rel_sc}

3D SC is a computer vision task aiming to plausibly fill in the holes and missing areas of a volume, predicting if they are free or occupied, based on the partially observed scene.
Oftentimes, this observation can be as little as a single depth image.
Similarly, as the extrapolation of this data inherently requires scene understanding \cite{palnet}, this task is oftentimes trained jointly as Semantic Scene Completion (SSC), additionally predicting semantic labels for the entire volume \cite{roldao20223d}.

To this end, a number of approaches have been proposed.
Song \etal~\cite{Song2017SemanticSC} introduced SSCNet, an end-to-end 3D Convolutional Neural Network (CNN) to complete occupancy and semantics of a volume from a single depth image. 
This was augmented by Li \etal~\cite{ddrnet} in DDRNet by additionally incorporating color data. 
They further address the large computation load of \cite{Song2017SemanticSC} through efficient decomposition of the 3D convolutions.
Liu \etal~\cite{palnet} introduce an architecture combining features of 2D and 3D representations of a depth image in PALNet. 
This is combined with a position aware loss, emphasizing boundaries and corners of the scene, to further improve performance. 
In CCPNet~\cite{Zhang2019CascadedCP}, Zhang \etal~utilize separated kernels for 3D convolutions to reduce the computation costs of 3D convolutions while maintaining input resolution. 
In contrast to earlier approaches \cite{Song2017SemanticSC, ddrnet, palnet} that combine features in serial or parallel fashion, CCPNet proposes a cascaded multi-scale feature aggregation scheme.
A more complex multistage approach is found in 3D-Sketch \cite{3dsketch}, integrating sparse geometry completion into SSC. 
While it offers marginally increased performance, complexity increases significantly.
In ForkNet \cite{Wang2019ForkNetMV}, Wang \etal~use an encode-generate architecture, where an input depth image is encoded into a latent space, from which reconstruction, completion, and semantic branches fork off to generate the output.

A common limitation of these methods is their significant computational complexity, making it hard to apply them on mobile robots with constrained hardware and real-time requirements. 
An incremental 3D SSC framework is presented by Wu \etal~\cite{wu_scfusion_2020} in SCFusion.
During 3D reconstruction, sub-volumes of the map are extracted, completed using a lightweight 3D CNN, and consequently fused back into the map.
Similar to ours, a recent work by Fehr \etal~\cite{fehr2019predicting} and Popovic \etal~\cite{popovic_volumetric_2021} propose to learn 2D depth completion, which is incrementally fused into the robot's occupancy map to clear free space more quickly for navigation.
However, this is inherently still limited by occlusions and can not fully complete the scene.

Although SC has been proposed for robotic navigation tasks, there remains a lack of works addressing incremental scene completed mapping and planning. 
In \cite{wu_scfusion_2020}, a single probabilistic voxel map is maintained.
When fusing scene completed volumes back into the global map to ensure safety only occupied parts are considered and with a lower probability compared to observed measurements.
However, this can lead to performance detriments as the robot may be blocked by erroneous completions and potentially valuable free space information is neglected. 
Similarly, \cite{popovic_volumetric_2021} estimate the certainty of the depth completions which is used as probabilistic weight for occupancy fusion.
While this speeds up discovery of free space significantly, erroneous free space is added to the map potentially compromising robot safety. 
Furthermore, active perception and planning aspects are not yet considered.

In contrast, we propose a novel multi-layer mapping framework keeping all data disentangled. 
This representation is interpretable and can guarantee safety, efficiency, or trade-offs thereof for efficient planning.


\subsection{Learning for Robotic Exploration}
\label{sec:rel_learning}

The problem of learning to explore unknown environments has received little attention. 
The majority of works directly try to learn the next action from the robot state.
For this purpose, RL approaches are in theory the most general, but can be difficult to train \cite{henderson2018deep} and thus tend to rely on significant simplifications, such as restriction to planning in 2D.
Zhu \etal~\cite{zhu2018} use an actor-critic RL approach to select one of six sectors to be used for classical planning.
Similarly, Niroui \etal~\cite{niroui2019} extract frontiers and use RL to select between them. 
Chen \etal~\cite{chen2019self} used RL with a pre-defined set of actions for a robot with omnidirectional sensor and kinematics.

Another approach is to use supervised learning to imitate a conventional planner, which can be computationally cheaper \cite{andersson2017deep}. 
Reinhart \etal~\cite{reinhart2020} learn to imitate an exploration planner for tunnel environments, choosing from fixed trajectories in eight regions. 
Bai \etal~\cite{Bai2017} similarly use a discretized action space of fixed angles on a circle, reducing computational cost over exhaustively evaluating the gains for all such actions. 

Lastly, recent works propose to combine learning with classical planners. 
Hepp \etal~\cite{hepp_learnscore_2018} combine a sampling-based planner with a 3D CNN to estimate the information gain of each sample.
Schmid \etal~\cite{schmid2022fast} further also learn an underlying informed sampling distribution.

While directly predicting the robot action or information gain of a viewpoint implicitly learns a prior about the scene distribution, these methods tend to lack interpretability, safety, and maturity for application in 3D exploration.

\section{Problem Statement} 
\label{sec:problem}

In this paper, we address the problem of volumetric exploration with scene completion using voxel maps. 

\minisec{World Representation} 
We consider an area of interest, discretized as a 3D voxel grid, $V\subset \mathbb{R}^3$. Each voxel $v \in V$ has a state in the robot map $M$ at time $t$: $M_t(v): \mathbb{R}^3 \mapsto \mathbb{M} = \{m_o, m_f, m_u\}$, denoting occupied, free, and unknown space, respectively. Initially $M_0(v) = m_u\; \forall v \in V$. 
We denote $M^\ast(v): \mathbb{R}^3 \mapsto \{m_o, m_f\}$ the true state of each voxel, and $\hat{V}_t = \{ v \in V | M_t(v) \neq m_u \}$ the observed space at time $t$.

\minisec{Online Informative Path Planning}
At each time step $t$, the robot takes a decision based on its current map on where to move next. We denote this decision as the NBV pose $P_t: M_t \mapsto  \{x_t,y_t, z_t, \psi_t\} \in \mathbb{R}^3 \times SO(2)$, parametrized as position and yaw
\footnote{We choose this parametrization for our experiments on an MAV. Any other parametrization from $\mathbb{R}^2$ to complete $SE(3)$ is also permissible.}.
The goal of a planning mission can then be formulated as a constrained optimization problem:

\begin{equation}
    \label{eq:problem_objective}
    \begin{aligned}
        P_0^\ast,\dots, P_{N_P}^\ast = \underset{P_0,\dots,P_{N_P}}{\arg\max}\ O(M_{N_P}) \\
        \text{subject to } \sum_{i=0}^{N_P} T(P_i) \leq T_{max}
    \end{aligned}
\end{equation}
where $P_0, \dots, P_{N_P}$ denotes the sequence of chosen view points $P_i$, $O(M)$ is the objective function, $T(P_i)$ is a cost of each view, and $T_{max}$ is the cost budget. 
As mobile robots such as MAVs are oftentimes limited by their operation time, without loss of generality, we use the execution time as the cost, i.e. $T(P_i) = \exectime(P_i)-\exectime(P_{i-1})$. 
Plainly, this means maximizing the objective $O$ within a time limit $T_{max}$.

\minisec{Objectives for Exploration with Imperfect Data}
Assuming imperfect measurements, scene completion, and mapping algorithms, the standard exploration objective:

\begin{equation}
    \label{eq:problem_exploration}
    O_{\text{exp}}(M) = \sum_{v \in V}\mathbb{I}(M(v) \neq m_u)
\end{equation}
where $\mathbb{I}(\cdot)$ is the indicator function, is no longer an appropriate measure of success. 
As the priorities within an exploration mission are highly application dependent, we propose the following combination objectives:

\begin{equation}
    \label{eq:problem_coverage}
    O_{\text{cov}}(M) = \sum_{v \in V}\mathbb{I}(M(v) = M^\ast(v))
\end{equation}
The \emph{coverage} aims to maximize the total amount of \emph{true} information in the map.

\begin{equation}
    \label{eq:problem_accuracy}
    O_{\text{acc}}(M) = \frac{1}{|\hat{V}|} \sum_{v \in \hat{V}}\mathbb{I}(M(v) = M^\ast(v))
\end{equation}
The \emph{accuracy} measures the quality of the obtained map.

\begin{equation}
    \label{eq:problem_safety}
    O_{\text{safe}} = \mathbb{I}(V_{\text{robot}, t} \subseteq \{v \in V | M^\ast(v)=m_f\} \forall t)
\end{equation}
The \emph{safety} objective requires that the volume occupied by the robot $V_{\text{robot}, t}$ must be completely contained in free space at all times $t$.
Lastly, the desire for rapid exploration is controlled via the time budget $T_{max}$.
For a specific application, these objectives can be traded off as desired:

\begin{equation}
\label{eq:problem_tradeoff}
    O(M)= \alpha_1 O_{\text{cov}}(M) + \alpha_2 O_{\text{acc}}(M) + \alpha_3 O_{\text{safe}} 
\end{equation}

\section{Approach} 
\label{sec:approach}

\begin{figure}
\centering
\includegraphics[width=\linewidth]{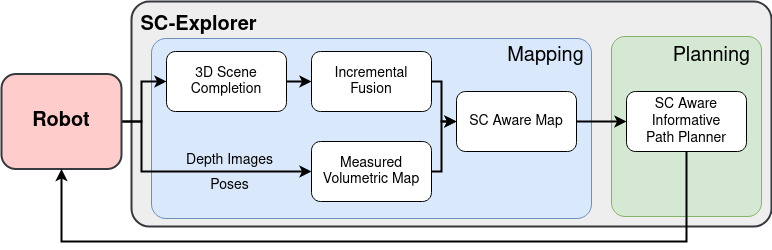}
\caption{System overview. Depth images are incrementally completed and fused together. The predicted map is combined with the fused measured data, leading to a hierarchical multi-layer map that can be used for SC aware planning.  }
\label{fig:overview}
\end{figure}

The central idea of our work is to incrementally complete the scene around the robot, filling in occlusions in the map and extrapolating into unknown space.
To this end, we continuously predict 3D scene completions based on the incoming depth images, fuse the predictions spatially over time, and combine them with the fused measured data into a hierarchical multi-layer map.
This map provides additional information for SC-aware Informative Path Planning (IPP) of the robot, or the end user. 
An overview of \emph{SC-Explorer}, our system for complete SC-aware mapping and planning, is shown in Fig.~\ref{fig:overview}.
Each component is further detailed below.


\subsection{3D Scene Completion}
\label{sec:approach_sc}

\begin{figure*}
\centering
\includegraphics[width=\textwidth]{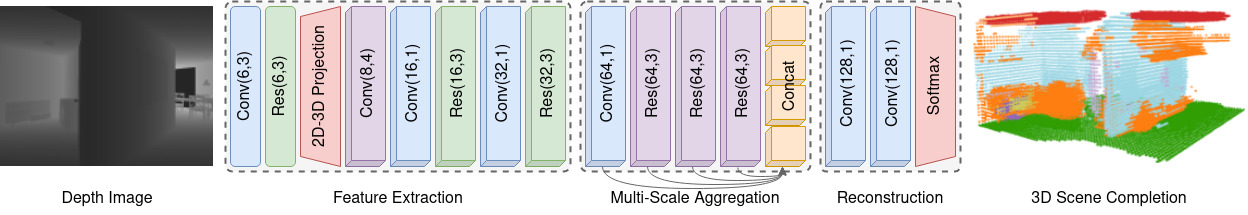}
\caption{3D scene completion network based on PALNet \cite{palnet}. 2D Layers are shown as planar blocks while 3D blocks are shown in 3D. We use Convolutional (Conv) and Residual convolutional  (Res) blocks, where BlockName(filters, kernel size) indicate the block parameters. Spatial reductions using dilated convolutions are shown in purple.}
\label{fig:network_architecture}
\end{figure*}

The first component in our pipeline is a 3D SC method. 
It is important to highlight that any SC approach can be substituted here, as our approach is agnostic with respect to the particular SC system used.
In this work, we use PALNet \cite{palnet}.
Due to the high computational complexity of the method, we base our network on the open-source implementation \cite{ssc-palnet-ddrnet-opensource} of PALNet and make some modifications to the feature extraction module.
An overview of the employed architecture is shown in Fig.~\ref{fig:network_architecture}.
In particular, we prune the 3D feature extraction branch, as the frame-wise flipped Truncated Signed Distance Field (TSDF) encoding followed by large 3D blocks are costly operations.
We further downscale the 3D processing part of the depth feature extractor by reducing the 3D volume with a dilated convolution layer.
To compensate for the reduced model capacity, we employ a slightly deeper architecture and increased number of extracted features. 
The network predicts grid of $60\times 60\times 36$ class probabilities, corresponding to a scene completed volume of $4.8\times 4.8\times \SI{2.88}{m}$ with a voxel size of $\nu=\SI{0.08}{m}$. 


\subsection{Incremental Fusion of Scene Completions}
\label{sec:approach_fusion}

In a second step, the obtained SC predictions are incrementally fused into the map. 
In this step, we do not yet consider the combination of SC with sensor measurements, but focus on how individual predictions can be fused together.
To this end, a probabilistic fusion approach is employed.
Each voxel $v$ in our SC-map accounts for the current SC occupancy probability $p_t(v)$.
As a prior $p_0(v)=0.5$ $\forall v$.
We follow the fully Bayesian approach of \cite{loop2016closed} to estimate $p_t(v)$:

\begin{equation}
\label{eq:fusion_bayes}
    p_t(v) = \frac{\prod_{i=1}^t \hat{p}_i(v)}{\prod_{i=1}^t \hat{p}_i(v) + \prod_{i=1}^t (1-\hat{p}_i(v))}
\end{equation}
where $\hat{p}_t(v)$ denotes the predicted occupancy probability at time $t$. Eq.~(\ref{eq:fusion_bayes}) can be reformulated for cumulative updates:

\begin{equation}
\label{eq:fusion_incremental}
    p_t(v) = \frac{p_{t-1}(v)\hat{p}_t(v)}{p_{t-1}(v)\hat{p}_t(v) + (1-p_{t-1})(1-\hat{p}_{t})}
\end{equation}
Adopting the log-odds formulation $l \coloneqq \log \left(\frac{p}{1-p}\right)$ results in an efficient incremental update rule:

\begin{equation}
\label{eq:fusion_logodds}
    l_t(v) = l_{t-1}(v) + \hat{l}_t(v)
\end{equation}
and allows storing a single value $l_t(v)$ in each voxel.

However, directly using the predicted occupancy confidence for $\hat{p}_t(v)$ can lead to inaccurate estimates, as these can be overconfident \cite{mccormac_fusion_2018, strecke_emfusion_2019, schmid_panoptic_2022}.
Furthermore, in the case of 3D scene completion, objects that have not been recognized correctly by the network or are completely unseen are oftentimes not captured in the prediction or predicted to be free space.
For this reason, we propose to model the incremental fusion of 3D SCs as an occupancy detection problem rather than a binary classification problem.

We therefore propose the \emph{occupancy} fusion update weight $\hat{l}^{occ}_t(v)$ as:

\begin{equation}
\label{eq:fusion_occupancy}
    \hat{l}^{occ}_t(v) = \left\{
    \begin{aligned}
    & \log \left(\frac{\bar{p}_f}{1-\bar{p}_f}\right), && \text{if predicted free} \\
    & \log \left(\frac{1 + \bar{p}(c(v,t))}{1-\bar{p}(c(v,t))}\right), && \text{otherwise}
    \end{aligned} \right .
\end{equation}
A constant low free space probability $\bar{p}_f$ is employed to incrementally refine the occupancy predictions, without overruling previous detections with missed or erroneous predictions. 
For this work, we choose $\bar{p}_f = 0.49$.

The second term of Eq.~(\ref{eq:fusion_occupancy}) models the occupancy probability for occupied predictions, with two main modifications.
First, instead of using the network confidence, a constant empirical estimate of the occupancy confidence $\bar{p}$ is employed.
We calibrate $\bar{p}$ individually for every class $c$ on a hold-out validation set of the training data, by computing the fraction of predicted voxels of each class that are occupied.
In our case, this results in highest confidence for very distinct objects ($\bar{p}(\text{sofa})=0.56$), medium confidence for the background ($\bar{p}(\text{floor})=0.41$), and low confidence for complex and diverse objects ($\bar{p}(\text{furniture})=0.3$).

In addition, to model the correct detection of objects, the second term of Eq.~(\ref{eq:fusion_occupancy}) allows only positive increments on the occupancy probability for any value of $\bar{p}$.
The term becomes identical to the original log-odds formulation $l'$ by applying the transformation $p' = \frac{1+p}{2}$.


\subsection{3D Scene Completed Mapping}
\label{sec:approach_mapping}

A major question is how the predicted SC should be combined with the sensor measurements.
Current methods primarily make use of occupancy maps to integrate the sensor measurements and the completions into a single grid \cite{wu_scfusion_2020,popovic_volumetric_2021}.
This has the advantage that no extra memory is consumed, but comes at the cost of not knowing which parts were \emph{predicted} by the network and which areas were \emph{measured} by the sensor. 
As this extra information can be highly valuable for mapping and planning, and crucial for safety when also free space is fused into the map, we instead propose a \emph{hierarchical} multi-layer mapping architecture.

Our first layer models only the sensor measurements, which are incrementally fused.
Although any algorithm is amenable, we use a TSDF map based on voxblox \cite{Oleynikova2017VoxbloxI3}. 
Consequently, the second layer consists of the \emph{SC layer} introduced in Sec.~\ref{sec:approach_fusion}.
We adapt the architecture of \cite{Oleynikova2017VoxbloxI3} also for the SC layer and incrementally allocate voxel blocks when new predictions arrive.
Notably, both layers are grid aligned with identical voxel sizes $\nu$, such that blocks and voxels have a one-to-one correspondence and can be efficiently queried.
Similarly, since our SC voxels only store one additional float, the memory consumption of our multi-layer map is only 33\% higher than that of the original TSDF map.

We argue that this is a worthwhile trade-off, as this multi-layer architecture enables valuable extra information for map look-ups.
These include that the measured map can never be impacted or corrupted by network predictions, potentially compromising the robot safety.
Second, this extra information can be used relevant for different planning schemes, which are discussed in more detail in Sec.~\ref{sec:approach_sc_planning} and Sec.~\ref{sec:approach_gains}.
Lastly, it allows discounting the predicted map with an additional confidence threshold $\tau_c\in[0, 1]$, without affecting the measured parts of the scene.

\begin{figure}
\centering
\includegraphics[width=0.9\linewidth]{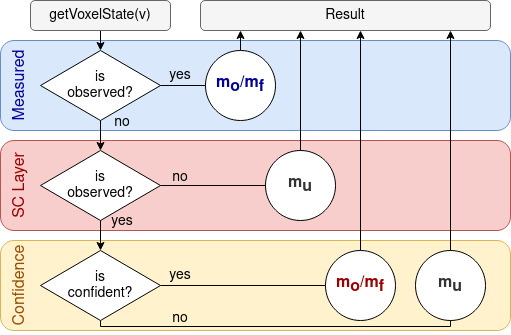}
\caption{Hierarchical map look-ups in our multi-layer map.}
\label{fig:map_lookup}
\end{figure}

In particular, we define the confidence cut-offs as:
\begin{equation}
    \label{eq:confidence}
    l_o = \log \left(\frac{1 + \tau_c}{1-\tau_c}\right), \quad l_f = \log \left(\frac{1 - \tau_c}{1 + \tau_c}\right)
\end{equation}
where voxels are considered occupied if $l_t(v) \geq l_o$ and considered free if $l_t(v) \leq l_f$, respectively.
The threshold $\tau_c \in [0, 1]$ can thus be chosen for a specific application very intuitively, as for $\tau_c=0$ voxels are occupied if $p_t(v) \geq 0.5$, and otherwise free.
The threshold extends linearly to voxels only being counted as occupied for $p_t(v)=1$ and free for $p_t(v)=0$ at $\tau_c=1$.
Note that voxels with $l_f < l_t(v) < l_o$ are considered to be unknown $M(v)=m_u$.
The proposed hierarchical map look-up is visualized in Fig.~\ref{fig:map_lookup}.

How SC is modeled in the map has a notable impact on the exploration objectives.
First, any addition of non-zero accuracy SC will add some true information to the map, increasing $O_{cov}(M)$ of Eq.~(\ref{eq:problem_coverage}).
However, it also has the potential to add false information, reducing $O_{acc}(M)$ shown in Eq.~(\ref{eq:problem_accuracy}).
A notable strength of our approach is that through the explicit representation of predicted areas and the added confidence, these objectives can be traded-off based on the application or the SC can be employed for planning only.

\subsection{Informative Path Planning}

Our IPP approach is largely based on the method described in \cite{Schmid20ActivePlanning}.
We adapt the planning architecture of \cite{Schmid20ActivePlanning} and briefly recapitulate the relevant features here, whereas we innovate primarily on planning mechanisms enabled by our multi-layer map in Sec.~\ref{sec:approach_sc_planning} and various information gain formulations in Sec.~\ref{sec:approach_gains}.

We employ a sampling-based NBV planning approach. 
While the robot moves towards the current NBV, a tree of Nodes $\{N_i\}$ is continuously expanded.
Each node $N_i=\{g(N_i),c(N_i),u(N_i)\}$ contains an information gain $g$, and associated cost $c$, and a final utility $u$.
The information gain $g$ estimates the positive impact of visiting a node $N_i$.
The cost estimates the negative effects of visiting a node $N_i$.
Lastly, the utility $u$ combines the gains and costs into a unified optimization target.

In this work, we employ the execution time of a segment as its cost by estimating the traversal time using velocity ramp dynamics for both position and orientation.
We further employ the utility formulation of \cite{Schmid20ActivePlanning}:
\begin{equation}
    u(N_i) =
        \max_{N_j \in \subtree(N_i)}
        \frac{
            \sum_{N_k \in \treepath(N_j)}g(N_k)
        }{
            \sum_{N_k \in \treepath(N_j)}c(N_k)
    }
    \label{eq:global_utility}
\end{equation}
where $\subtree(N_i)$ indicates all nodes connected to $N_i$ and subsequent children, and $\treepath(N_i)$ is the set of nodes connecting $N_i$ to the root of the tree.
The tree is continuously rewired to maximize $u$ anywhere in the tree.
More plainly, the algorithm thus identifies at each time step a sequence of nodes that maximizes the total accumulated gain divided by the total accumulated cost.
Once the NBV is reached, the first node of the currently best path is executed and the tree is updated.


\subsection{Planning on Scene Completed Maps}
\label{sec:approach_sc_planning}

In the presented planning approach, the map is queried in three cases.
Due to the capability of our multi-layer map to explicitly represent measured and predicted areas, each of these can be handled differently. 

The first case considers collision checking when drawing new samples, expanding, or updating the tree.
Here we differentiate between two modes, \emph{conservative} collision checking treats only \emph{measured} voxels as traversable.
This guarantees that the robot will always stay as safe as if no SC were used, even if free space data is fused in the SC-map, maximizing $O_{safe}$ of Eq.~(\ref{eq:problem_safety}).
Alternatively, \emph{optimistic} planners can plan robot paths into areas that are only predicted to be free.
This can potentially speed up exploration progress by allowing the robot more space for planning, but safety of the robot can no longer be guaranteed, thus trading-off $O_{safe}$ of against $O_{cov}(M)$ in Eq.~(\ref{eq:problem_tradeoff}) for specific applications.

Second, we compute the information gain $g(N_i)$ using sensor forward simulation:

\begin{equation}
    g(N_i) = \sum_{v\in\visible(P(N_i))} I(v) 
    \label{eq:ray_casting}
\end{equation}
where $\visible(P)$ is the set of all voxels that are visible from pose $P$, $P(N_i)$ is the NBV pose of node $N_i$, and $I(v)$ is the information contribution of each individual voxel.
We employ orientation optimization as in \cite{Schmid20ActivePlanning, history_nbvp, Selin_nbv_fron} to select the yaw $\psi$ of each position sample:
\begin{equation}
    \psi_{N_i} = \arg\max_\psi g(x_{N_i}, y_{N_i},z_{N_i},\psi)
    \label{eq:yaw_optimization}
\end{equation}
These can be optimized greedily per view point, as nodes $N_i$ are assumed to be independent in \cite{Schmid20ActivePlanning}, and the best yaw for each pose thus automatically maximizes Eq.~(\ref{eq:global_utility}).

The visible voxels $\visible(P)$ are computed using iterative ray-casting \cite{Schmid20ActivePlanning}, where again there are two options how the map can be employed, depending on whether the SC layer is considered during ray-casting.
We term the method \emph{SC-blocking} if rays are terminated on voxels that are predicted to be occupied.
This has the advantage, that if occupied voxels are predicted with high quality, $\visible(P)$ more closely reflects the volume that in reality will be observed, leading to a more accurate gain estimate.
However, if there are wrong predictions, these can block out simulated sensor views and hinder exploration progress.
In contrast, \emph{non-blocking} ray-casting has neither of the advantages nor disadvantages, and relies purely on the measured voxels to estimate $\visible(P)$. 

Lastly, the map is queried when computing the information contribution of a voxel $I(v)$, which can take advantage of the proposed map in various forms, discussed in more detail in Sec.~\ref{sec:approach_gains}.


\subsection{Information Gain Formulations}
\label{sec:approach_gains}

The information gain is computed following Eq.~(\ref{eq:ray_casting}), using different formulations for $I(v)$.
For clearer notation, we denote as $\mathbb{S}$ the set of all voxels \emph{measured} by the sensor, and as $\mathbb{P}$ those voxels that are \emph{predicted} by the SC but are not measured.
Using this formulation, the classic volumetric exploration gain \cite{Bircher2016RecedingH, Schmid20ActivePlanning} aiming to measure as much unmeasured volume as possible can be defined as:

\begin{equation}
\label{eq:gain_exploration}
I_{exp}(v) = \begin{cases}
0, & \text{if } v\in \mathbb{S} \\
1, & \text{otherwise}
\end{cases}
\end{equation}

As our proposed map explicitly models measured and predicted space, this information can be leveraged for SC-aware gain formulations.
The central idea of our proposed gain $I_{sc}(v)$ is based on two assumptions.
First, the network predicts what the unmeasured scene around the robot will look like, based on what is partially observed.
Second, for exploration it is sufficient to plan locally, as the majority of relevant information guiding future steps will only be uncovered as the robot moves \cite{schmid2021glocal}.
Combining these two parts, if the network is reasonably accurate, the predicted but not measured parts are exactly what the robot ought to observe, and has sufficient extent to lead to efficient exploration paths.
We thus propose:

\begin{equation}
\label{eq:gain_sc}
I_{sc}(v) = \begin{cases}
1, & \text{if } v\in \mathbb{P} \\
0, & \text{otherwise}
\end{cases}
\end{equation}
This gain formulation has several advantages, as it leads to the robot implicitly adapting to the type of environment.
For example, in straight corridor the robot tends to move straight in the center as this observes most of a predicted straight corridor.
The robot thus chose a close to optimal path without 'knowing' that it is in a corridor.

We further explore several variations of $I_{sc}(v)$. 
First, we propose a hybrid gain, combining Eq.~(\ref{eq:gain_exploration}) and Eq.~(\ref{eq:gain_sc}):

\begin{equation}
\label{eq:gain_hybrid}
I_{hybrid}(v) = I_{exp}(v) + I_{sc}(v) \begin{cases}
2, & \text{if } v\in \mathbb{P} \\
0, & \text{if } v\in \mathbb{S} \\
1, & \text{otherwise}
\end{cases}
\end{equation}
The hybrid gain still emphasizes the predicted areas, but retains a pure exploration aspect for areas beyond the extent of, or not captured by, the prediction.
Since the proposed SC fusion mechanism emphasizes on detection of objects, we further investigate a variant of $I_{sc}(v)$ focusing on observing as much surface as possible:

\begin{equation}
\label{eq:gain_occ}
I_{occ}(v) = \begin{cases}
1, & \text{if } v\in \mathbb{P} \wedge M_t(v) = m_o\\
0, & \text{otherwise}
\end{cases}
\end{equation}
Lastly, we explore weighting $I_{sc}(v)$ by the confidence of $v$ in the SC layer, encouraging the planner to prioritize high certainty predictions.
We therefore employ a linear weight on the probability $p_t(v)$:

\begin{equation}
\label{eq:gain_conf}
I_{conf}(v) = \begin{cases}
|0.5 - p_t(v)|, & \text{if } v\in \mathbb{P}\\
0, & \text{otherwise}
\end{cases}
\end{equation}
Note that since Eq.~(\ref{eq:global_utility}) performs comparative optimization, scaling of the different gains is not required.




\section{Experimental Setup}
\label{sec:experiment_setup}

\subsection{Learning Setup}

The employed network is trained using the setup of \cite{ssc-palnet-ddrnet-opensource} on the NYU dataset \cite{nyu}, using the 3D ground-truth annotations of \cite{guo2015predicting}. 
The network is trained for 50 epochs with an initial learning rate of 0.01 and exponential decay of 0.1 every 10 epochs. 
Although we primarily use the occupancy probability of the SC, we train the network also with the semantic loss as these can aid in improving the geometric predictions \cite{palnet}.


\subsection{Simulation Environment}
\begin{figure}
\centering
\includegraphics[width=0.5\textwidth]{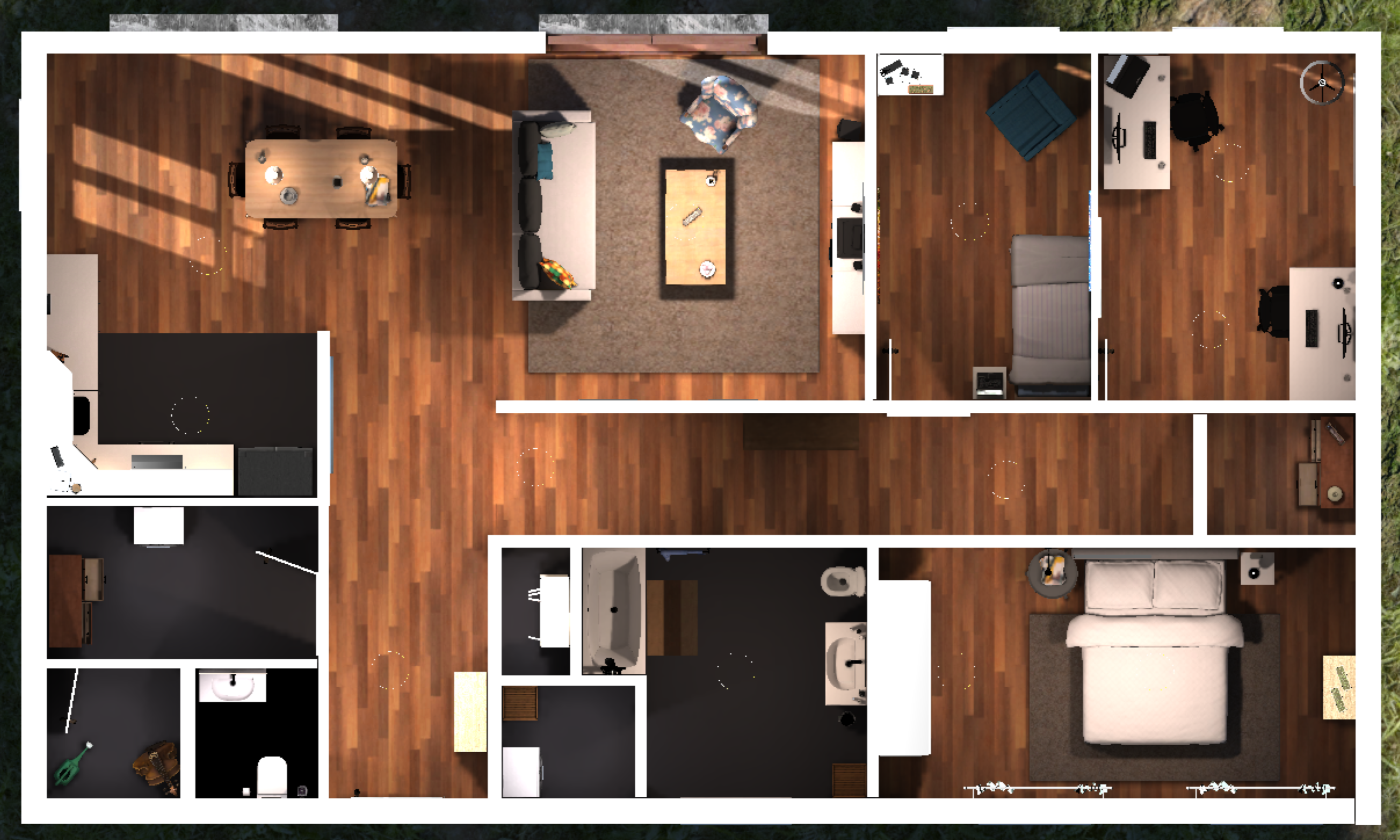}
\caption{High fidelity simulation environment combining the three scenes of \cite{zurbrugg_embodied_2022}. Challenges include the complex room layout, the large extent, as well as the detailed and distributed furniture.}
\label{fig:simulation}
\end{figure}

Since in active planning problems the robot can act based on the environment, a simulator rather than a dataset is required. 
To thoroughly evaluate the proposed methods, a high fidelity simulator and complex environment is employed.
We use the simulation setup of \cite{schmid2021glocal}, combining the photo realistic rendering capabilities of Unreal Engine 4\footnote{\url{https://www.unrealengine.com/en-US/}} and simulation interfaces of Microsoft AirSim \cite{airsim2017fsr}.
All approaches are tested in a complex and extensive indoor environment shown in Fig.~\ref{fig:simulation}, combining the three scenes of \cite{zurbrugg_embodied_2022}.
To allow for a fair comparison of the impact of 3D SC on mapping and planning, ground truth poses and sensor measurements are used.
Nonetheless, the presented method can readily be employed with imperfect estimates and measurements.
Lastly, we deploy the network solely pre-trained on the real depth images of \cite{nyu}.
Its application with the simulator therefore represents a domain shift, as can be expected when deploying the system in the real world.
For evaluation, a ground truth map of all \emph{observable} space is extracted from the simulator, containing all free space connected to the initial position and extending one voxel into occupied areas, thus, e.g. the inside of walls or objects are not evaluated.  

\subsection{Evaluation Metrics}

To accurately evaluate the performance of each approach the following set of metrics is employed.

\minisec{Coverage Metrics} To assess the extent to which the environment is represented in the robot map, we use \emph{exploration} $\mathcal{E}$ and \emph{coverage} $\mathcal{C}$, defined by the right hand terms of Eq.~(\ref{eq:problem_exploration}) and Eq.~(\ref{eq:problem_coverage}), to measure the amount of total information and the amount of true information in the map, respectively.
We further denote $\mathcal{M}$ the \emph{measured} volume, i.e. the exploration when only voxels measured by the sensor are counted.

\minisec{Quality Metrics} These metrics assess the quality of the robot map. 
The \emph{overall precision} $\mathcal{P}$ follows the definition in Eq.~(\ref{eq:problem_accuracy}) and represents the fraction of information in the map that is true. 
For detailed assessment, the \emph{precision} of \emph{free} $\mathcal{P}_f$ and \emph{occupied} space $\mathcal{P}_o$ is computed as:

\begin{equation}
    \label{eq:metric_precision_fo}
    \mathcal{P}_{(\cdot)} = \frac{\sum_{v \in V}\mathbb{I}(M(v) = m_{(\cdot)} \wedge M^\ast(v) = m_{(\cdot)})}{\sum_{v \in V}\mathbb{I}(M(v) = m_{(\cdot)} )}
\end{equation}
Similarly, the \emph{recall} of \emph{free} $\mathcal{R}_f$ and \emph{occupied} space $\mathcal{R}_o$ is computed as:

\begin{equation}
    \label{eq:metric_recall_fo}
    \mathcal{R}_{(\cdot)} = \frac{\sum_{v \in V}\mathbb{I}(M(v) = m_{(\cdot)} \wedge M^\ast(v) = m_{(\cdot)})}{\sum_{v \in V}\mathbb{I}(M^\ast(v) = m_{(\cdot)} \wedge M(v) \neq m_u)}
\end{equation}
and represents the amount of actually occupied/free voxels in the map that were correctly captured as such. 

Notably, the quantities $\mathcal{R}_o$ and $\mathcal{P}_f$ are related and safety critical, as the navigation relies on the fact that areas denoted free in the map are not actually occupied.

\minisec{Performance Metrics} To asses the effectiveness with respect to the cost constraint of Eq.~(\ref{eq:problem_objective}) of any metric $f(M)$, the \emph{final} performance $F_f = f(M_{T_{max}})$ is evaluated.
To account for varying budgets $T_{max}$, we also compute \emph{time} $\mathcal{T}_{\bar{f}}$ till a goal performance $\bar{f}$ is reached:

\begin{equation}
    \label{eq:metric_time}
    \mathcal{T}_{\bar{f}} = \sum_{i=0}^{N_P} T(P_i) \cdot \mathbb{I}(f(M_i) < \bar{f})
\end{equation}
as well as the \emph{expected} performance $E_f$ if $T_{max}$ was varied within a range of possible budgets $[t_{min},t_{max}]$:

\begin{equation}
    \label{eq:metric_expected}
    E_f = \underset{T_{max}\sim\mathcal{U}(t)}{\mathbb{E}[F_f]} = \frac{1}{t_{max}-t_{min}} \int_{t_{min}}^{t_{max}} F_f\ \partial T_{max}
\end{equation}
Since the employed sampling-based planner is stochastic, each experiment is repeated 10 times and the means and standard variations are reported.


\subsection{Employed Hardware and Parameters}
\label{sec:hardware}
To guarantee the applicability of the approach for a payload constrained autonomous robot, all experiments were conducted using only portable hardware.
The 3D SC network is run on a NVIDIA Jetson NX with 8 GB of memory.
The mapping and planning are executed on an Intel NUC with an i7-8650U CPU @ 1.90GHz, where 1 thread is used for planning and $\leq$1 thread is used for mapping.

Parameters for the simulated robot and mapping and planning system are identical throughout all experiments and listed in Tab.~\ref{tab:params}.

\begin{table}
    \centering
    \caption{System Parameters.}
    \begin{adjustbox}{max width=\columnwidth}
    \begin{tabular}{lcc|lcc}
        Parameter & & Value & Parameter & & Value \\
    \midrule
        Voxel Size & $\nu$ & $\SI{0.08}{\meter}$ & Velocity & $v_{max}$ & $\SI{1}{\meter\per\second}$\\
        Sensing Range & $r_s$ & $\SI{5}{\meter}$ & Yaw Rate & $\dot{\psi}_{max}$ & $\SI{90}{\degree\per\second}$ \\
        Field of View & FoV & $\SI{90}\times\SI{73.7}{\degree}$ & Collision Radius & $r_c$ & \SI{0.35}{\meter} \\
    \bottomrule
    \end{tabular}
    \end{adjustbox}
    \label{tab:params}
\end{table}


\section{Evaluations}
\label{sec:results}

\subsection{3D Scene Completion}
\label{sec:result_network}

\begin{table}
    \centering
    \caption{Occupancy Prediction on NYU \cite{nyu}.}
    \begin{adjustbox}{max width=\columnwidth}
    \begin{tabular}{lcccc}
        Method & Precision $\mathcal{P}_o$ [\%] & Recall $\mathcal{R}_o$ [\%] & IoU [\%] & FPS [Hz] \\
    \midrule
        PALNet (Ours) & \textbf{56.8} & 93.4 & \textbf{54.4} & \textbf{1.24} \\
        PALNet (Original) & 56.5 & \textbf{93.9} & \textbf{54.4} & 0.66 \\
    \midrule
        \multicolumn{5}{c}{{\footnotesize Highest number shown in bold.}}
    \end{tabular}
    \end{adjustbox}
    \label{tab:network}
\end{table}

To evaluate the effectiveness of our network modifications, we compare it to the original PALNet \cite{palnet} as provided in \cite{ssc-palnet-ddrnet-opensource}. 
We evaluate the capacity to correctly predict the occupancy of each voxel on the NYU test set \cite{nyu}. Precision, recall, Intersection over Union (IoU), and frame rate (FPS) are reported in Tab.~\ref{tab:network}.
We observe that the modified feature extractor is able to produce scene completions of very similar quality to the original approach. 
However, due to reduced model complexity, the inference rate on a portable NVIDIA Jetson is almost doubled, now achieving interactive rates $>\SI{1}{Hz}$ which is similar to the planning update rates.
We also note that the re-produced performance using \cite{ssc-palnet-ddrnet-opensource} is a bit lower than that originally reported in \cite{palnet}.
Nonetheless, as stated in Sec.~\ref{sec:approach_sc}, the quality of the scene completion is not the main focus of this work, and we will show that the achieved performance is sufficient to facilitate mapping and planning on a mobile robot.


\subsection{Incremental Fusion of Scene Completions}
\label{sec:results_fusion}

\begin{figure*}
\centering
\begin{subfigure}{.33\textwidth}
  \centering
  \includegraphics[width=\linewidth]{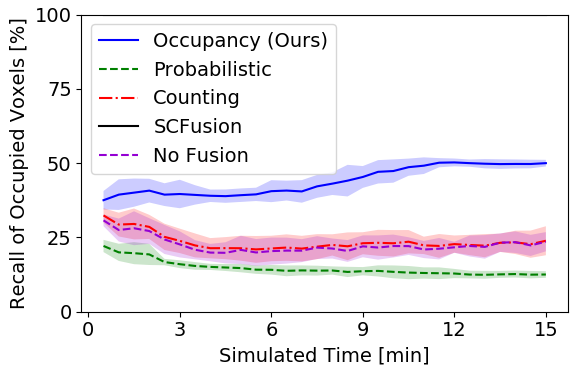}
\end{subfigure}%
\begin{subfigure}{.33\textwidth}
  \centering
  \includegraphics[width=\linewidth]{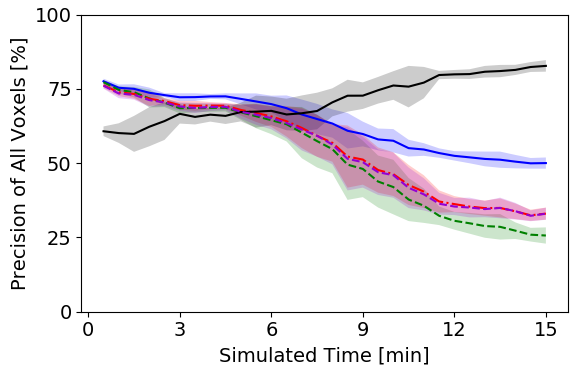}
\end{subfigure}%
\begin{subfigure}{.33\textwidth}
  \centering
  \includegraphics[width=\linewidth]{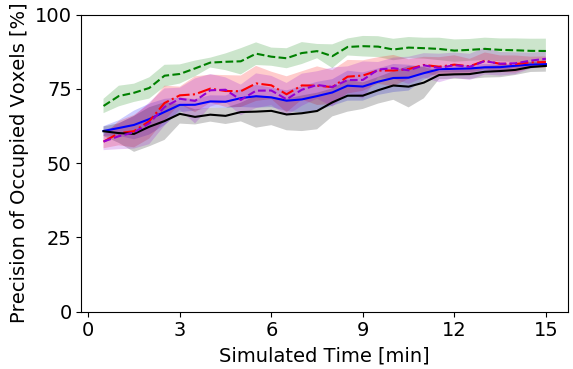}
\end{subfigure}%
\caption{Quality metrics recall $\mathcal{R}_o$ (left), overall precision $\mathcal{P}$ (center), and occupied precision $\mathcal{P}_o$ (right) for various fusion strategies over time. Each method receives identical scene completions as input. Only voxels that are not observed by the sensor are evaluated. }
\label{fig:fusion_quality}
\end{figure*}

\begin{figure}
\centering
\includegraphics[width=\linewidth]{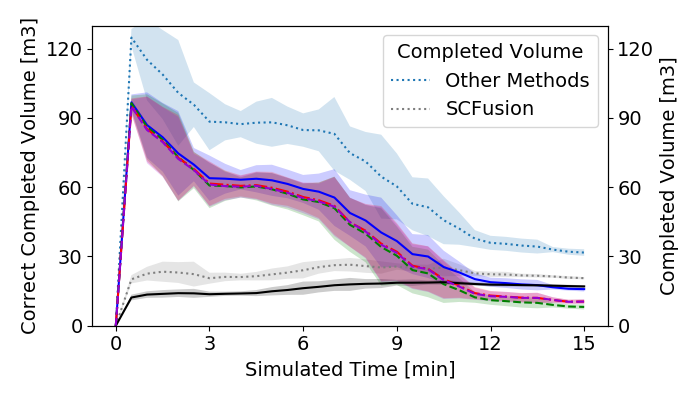}
\caption{Coverage $\mathcal{C}$ of the fusion methods listed in Fig.~\ref{fig:fusion_quality} (same colors, left scale) and their exploration $\mathcal{E}$ (in dotted, right scale).}
\label{fig:fusion_coverage}
\end{figure}

We evaluate different fusion methods by collecting the SC predictions for 10 runs of a classical SC-unaware exploration planner, such that the exact same input predictions are fused by every approach.
In order to meaningfully assess the utility of the fused SC-maps, only voxels that are not observed by the sensor are evaluated.

We compare our proposed \emph{Occupancy} fusion against the fully \emph{Probabilistic} approach \cite{popovic_volumetric_2021, mccormac_semanticfusion_2017} and estimates based on \emph{Counting} \cite{mccormac_fusion_2018, strecke_emfusion_2019, schmid_panoptic_2022}, which have found significant success in semantic mapping.
We further compare against the approach of \emph{SCFusion} \cite{wu_scfusion_2020}, which only fuses occupied predictions in areas not yet observed by the sensor, and a naive \emph{No Fusion} approach that simply overwrites the previous measurements.
The resulting quality and coverage metrics are shown in Fig.~\ref{fig:fusion_quality} and Fig.~\ref{fig:fusion_coverage}, respectively.

First, we observe that the initial network performance, most notably the recall $\mathcal{R}_o$, in this new environment is lower than in Tab.~\ref{tab:network}.
This can be explained by the domain shift from the similar train and test splits of NYU \cite{nyu} to our high fidelity simulator.
Similar behavior can be expected when applied to other environments, suggesting that our simulator represents an adequate evaluation setting. 

Fig.~\ref{fig:fusion_quality} demonstrates the capacity of our method to capture and fuse detected objects, reflected in significantly improved $\mathcal{R}_o$.
Notably, it is the only method that improves $\mathcal{R}_o$ over time.
This is explained by our modeling assumption of detecting objects rather than classifying each voxel, as the predicted free space at not recognized objects is fused with high confidence and leads to a decrease in $\mathcal{R}_o$ over time for the others.
Note that SCFusion naturally has $\mathcal{R}_o=100\%$ as only occupied voxels are fused.

While our approach not only achieves high $\mathcal{R}_o$, it also significantly improves in overall precision $\mathcal{P}$ over the baselines.
While $\mathcal{P}$ tends to decrease for all methods as the environment becomes more explored, this effect is diminished for our approach.
The only exception is SCFusion, however its map only integrates occupied predictions.

Upon closer inspection of $\mathcal{P}_o$ (Fig.~\ref{fig:fusion_quality}, right), we observe that our approach slightly improves in $\mathcal{P}_o$ over only fusing occupied voxels, showing the ability of our method to further refine uncertain occupancy predictions.
We note that Probabilistic is very conservative, manifested in the high $\mathcal{P}_o$ but low $\mathcal{R}_o$.
Interestingly, Counting and No Fusion show very similar behavior, highlighting the importance of calibrated confidence estimates.
Furthermore, we observe that, although $\mathcal{P}_o$ is similar for Occupancy, Counting, and No Fusion, $\mathcal{P}$ accounting for occupied and free voxels is higher for our method.
This indicates that free space is also more accurately represented using our approach, although at a lower precision than occupied as $\mathcal{P}$ generally $< \mathcal{P}_o$.

This also shows in the coverage reported in Fig.~\ref{fig:fusion_coverage}.
While our approach and SCFusion converge to a similar coverage level eventually, including free space predictions makes the fusion task more challenging, but adds significant amounts of additional information earlier that can be used for robot planning.
This is similarly reflected in the total completed but not measured volume shown on the right scale, which decreases as exploration becomes more complete.


\subsection{Impact of the Confidence Threshold}

\begin{figure}
\centering
\includegraphics[width=\linewidth]{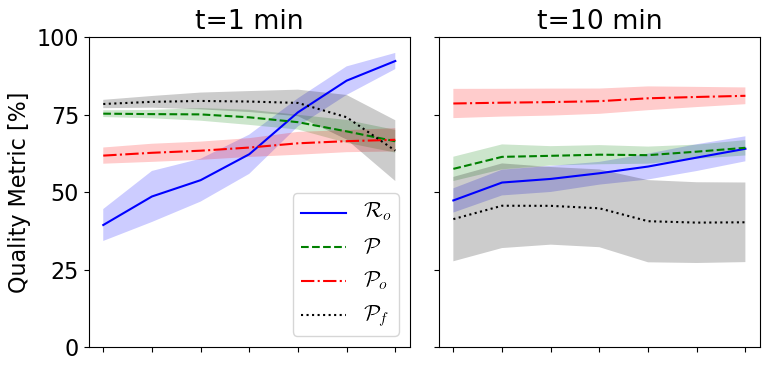}
\includegraphics[width=\linewidth]{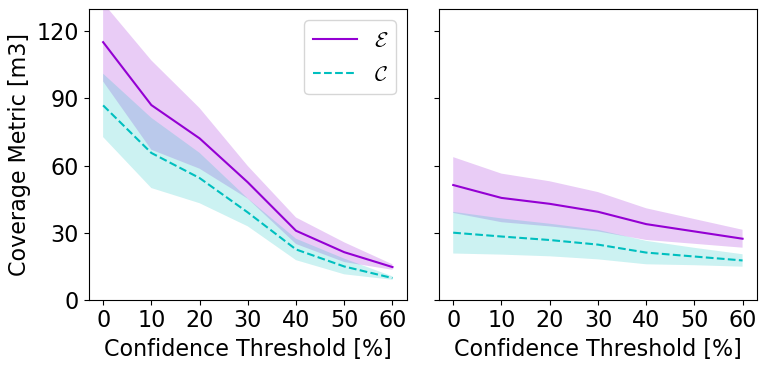}
\caption{Impact of the confidence threshold $\tau_c$ on quality metrics $\mathcal{R}_o$, $\mathcal{P}$, $\mathcal{P}_o$, and $\mathcal{P}_f$ (top) and coverage metrics $\mathcal{E}$ and $\mathcal{C}$ (bottom), early during exploration (left) and later on in the mission (right).}
\label{fig:confidence}
\end{figure}

The impact of different values for the confidence threshold $\tau_c$ is evaluated using the identical setup of Sec.~\ref{sec:results_fusion} and the proposed occupancy fusion mechanism. 
Fig.~\ref{fig:confidence} shows the quality metrics $\mathcal{R}_o$, $\mathcal{P}$, $\mathcal{P}_o$, and $\mathcal{P}_f$, as well as coverage metrics $\mathcal{E}$ and $\mathcal{C}$ for different values of $\tau_c$.
Early during exploration ($t=\SI{1}{min}$), comparably few predictions have been fused and the voxel confidences are thus generally lower. 
This is reflected in the strong impact of the confidence threshold on each metric.
As expected, a higher threshold results in higher quality. 
This shows primarily for the recall $\mathcal{R}_o$, and slightly in the precision $\mathcal{P}_o$. 
The precision of free voxels $\mathcal{P}_f$ stays unaffected, until it declines for high $\tau_c$.
This is due to the fact that, since in our fusion method free space is fused with constant low probability, there are barely any free space voxels that meet the confidence criterion for high $\tau_c$, manifesting in $\mathcal{P}\approx\mathcal{P}_o$. 
Similarly, the coverage metrics show that notable amounts of low-confidence information is discarded from the map.

Later during exploration ($t=\SI{10}{min}$) alike behaviors can be observed, although a bit less pronounced as confidences are generally higher.
This shows notably in the coverage, which is as in Sec.~\ref{sec:results_fusion} overall lower as more of the scene is measured, but fewer voxels are discarded as $\tau_c$ increases.
This effect also manifests in the quality metrics, where $\mathcal{P}_f$ is more stable and thus $\mathcal{P}$ also tends to increase.

These findings show that the additional confidence layer enabled by our hierarchical map representation can be an efficient tool to trade off coverage vs. accuracy in the areas that are added to the map by the SC.
This beneficial effect is more pronounced the more accurate the prediction uncertainty can be modeled.
We thus believe that more complex uncertainty estimators or better models of free space confidence for SC predictions is an interesting direction for future research.


\subsection{Scene Completed Mapping Performance}
\label{sec:results_map_type}

\begin{figure*}
\centering
\includegraphics[width=\linewidth]{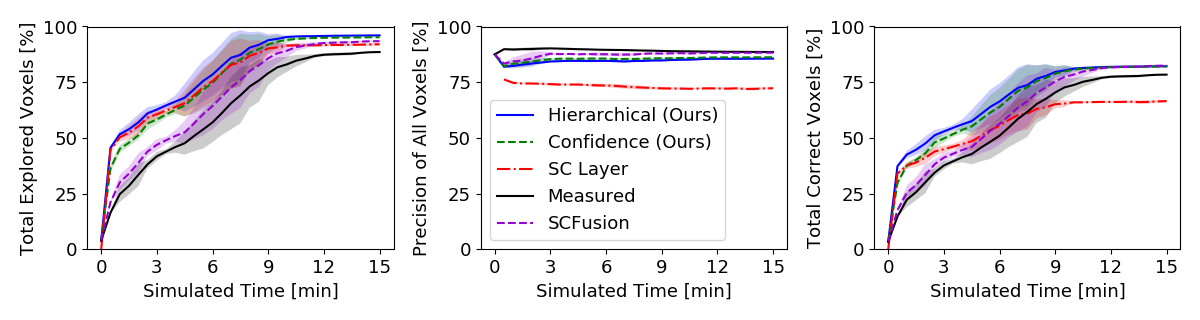}
\caption{Exploration $\mathcal{E}$ (left), precision $\mathcal{P}$ (center), and coverage $\mathcal{C}$ (right) for different maps based on identical sensor and SC inputs. }
\label{fig:map_type}
\end{figure*}

\begin{table}
    \centering
    \caption{Performance metrics for different mapping approaches.}
    \begin{adjustbox}{width=\linewidth}
    \begin{tabular}{lcccccc}
    Method / Metric [\%] & $F_\mathcal{E}$ & $E_\mathcal{E}$ & $F_\mathcal{P}$ & $E_\mathcal{P}$ & $F_\mathcal{C}$ & $E_\mathcal{C}$ \\
    \midrule
    Hierarchical (Ours)  & \textbf{95.9}  & \textbf{76.8} & 85.5 & 84.5 & \textbf{81.9} & \textbf{65.0}  \\
    Confidence (Ours) & 95.0 & 73.7 & 86.1 & 85.4 & 81.8 & 63.0 \\
    SC Layer & 91.9 & 73.9 & 72.2 & 73.2 & 66.3 & 54.0 \\
    Measured & 82.5 & 55.6 & \textbf{88.9} & \textbf{89.4} & 73.3 & 49.6 \\
    SCFusion  & 92.8 & 65.1 & 88.1 & 87.3 & 81.8 & 57.1 \\ 
    \midrule
    \multicolumn{7}{c}{\makecell{\footnotesize Average final $F_{(\cdot)}$ and expected $E_{(\cdot)}$ exploration $\mathcal{E}$, precision $\mathcal{P}$,\\ and coverage $\mathcal{C}$ performance. Highest number shown in bold.}}
    \end{tabular}
    \end{adjustbox}
    \label{tab:map_type}
\end{table}

To isolate the impact of the mapping approach, we evaluate numerous methods again on data collected from identical trajectories obtained by the pure exploration planner, and identical SC prediction generated by our network.
However, this time the entire map is considered for evaluation.
We compare our \emph{Hierarchical} approach detailed in Sec.~\ref{sec:approach_mapping} and our approach with a moderate \emph{Confidence} of $\tau_c=\num{10}$\% to several baselines.
First, we evaluate only using the \emph{SC Layer} (Sec.~\ref{sec:approach_mapping}) without explicit modeling of the sensor measurements. 
Nonetheless, note that the sensor data is implicitly captured as they are the input to the SC network and the predictions for the observed inputs tend to be of high quality.
Second, we evaluate only the \emph{Measured} map created by voxblox \cite{Oleynikova2017VoxbloxI3}, representing a traditional SC-unaware mapping approach.
Lastly, we compare against the full pipeline of \emph{SCFusion} \cite{wu_scfusion_2020}, adding occupied predictions to the yet unobserved map and integrating sensor data using probabilistic fusion. 
However, identical SC predictions as for the other methods are used as input to SCFusion to allow for a fair comparison.

Fig.~\ref{fig:map_type} shows the exploration $\mathcal{E}$, precision $\mathcal{P}$, and coverage $\mathcal{C}$ of each method.
Naturally, $\mathcal{E}$ is higher for the methods employing completions, as more data is added to the map through SC.
This is most pronounced early on, where the overlap of predicted and measured data is lower.
Notably, while SC Layer converges to a higher $\mathcal{E}$ than Measured, our hierarchical map representation combines the predictions in non-measured areas and the measured map in parts not detected by the network, leading to the most overall data in the map.

The highest precision $\mathcal{P}$ is achieved when only the measured data is considered.
However, we note that, even when using perfect poses and sensor measurements, $\mathcal{P}$ is only $\sim\num{90}$\% due to integration and discretization artefacts.
Similarly, the predicted-only SC Layer achieves a fairly high and consistent $\mathcal{P}\sim\num{75}$\%.
Combining both the measured and predicted data, our hierarchical approach starts with $\mathcal{P}$ in between measured and SC layer, but is able to continuously improve its map quality as more predictions and measurements are fused.

Both these aspects combine in the total true information $\mathcal{C}$.
Here we observe that, due to the notably higher accuracy of the data, Measured converges to a higher level than SC Layer.
Nonetheless, Hierarchical is able to combine true information from each modality, resulting in the most true information $\mathcal{C}$, both early on and after convergence.

Confidence and SCFusion both fall within the spectrum between Measured and Hierarchical. 
They show similar behavior in the coverage vs. quality trade-off, where SCFusion is the more conservative approach, adding a bit less information but achieving slightly higher $\mathcal{P}$.

The data is further summarized in the final $F_{(\cdot)}$ and expected $E_{(\cdot)}$ performance shown in Tab.~\ref{tab:map_type}.
Again, the precision $\mathcal{P}$ mostly separates between SC Layer and the other methods, but is comparably stable as $F_{\mathcal{P}}\sim E_{\mathcal{P}}$.
While the coverage $F_{\mathcal{C}}$ converges to similar values for all hybrid methods, $E_{\mathcal{C}}$ highlights the capability of our method to quickly cover an environment if the time $T_{max}$ is of the essence, such as in disaster response scenarios.


\subsection{Exploration on Scene Completed Maps}

\begin{figure}
\centering
\includegraphics[width=0.49\linewidth]{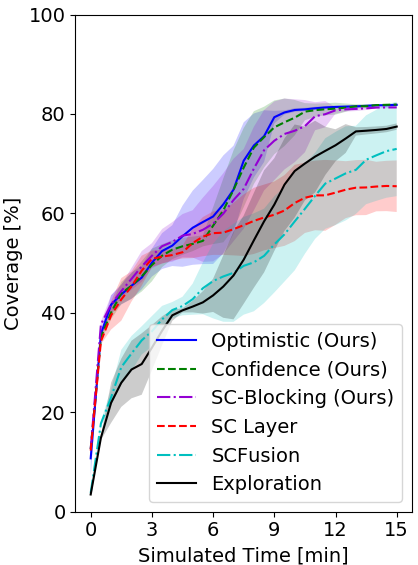}
\includegraphics[width=0.49\linewidth]{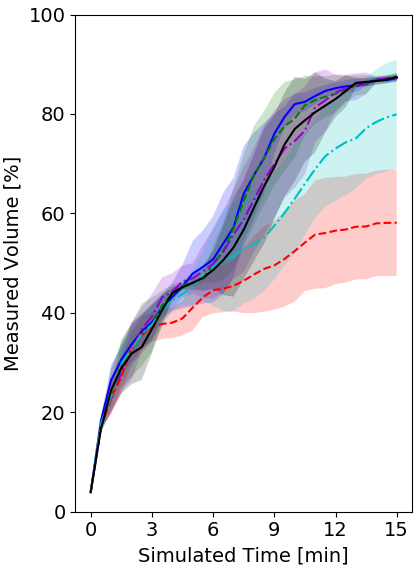}
\caption{Coverage $\mathcal{C}$ (left) and measured volume $\mathcal{M}$ (right) over time for a classical exploration planner using different mapping methods. }
\label{fig:map_exp}
\end{figure}

\begin{figure}
\centering
\includegraphics[width=0.49\linewidth]{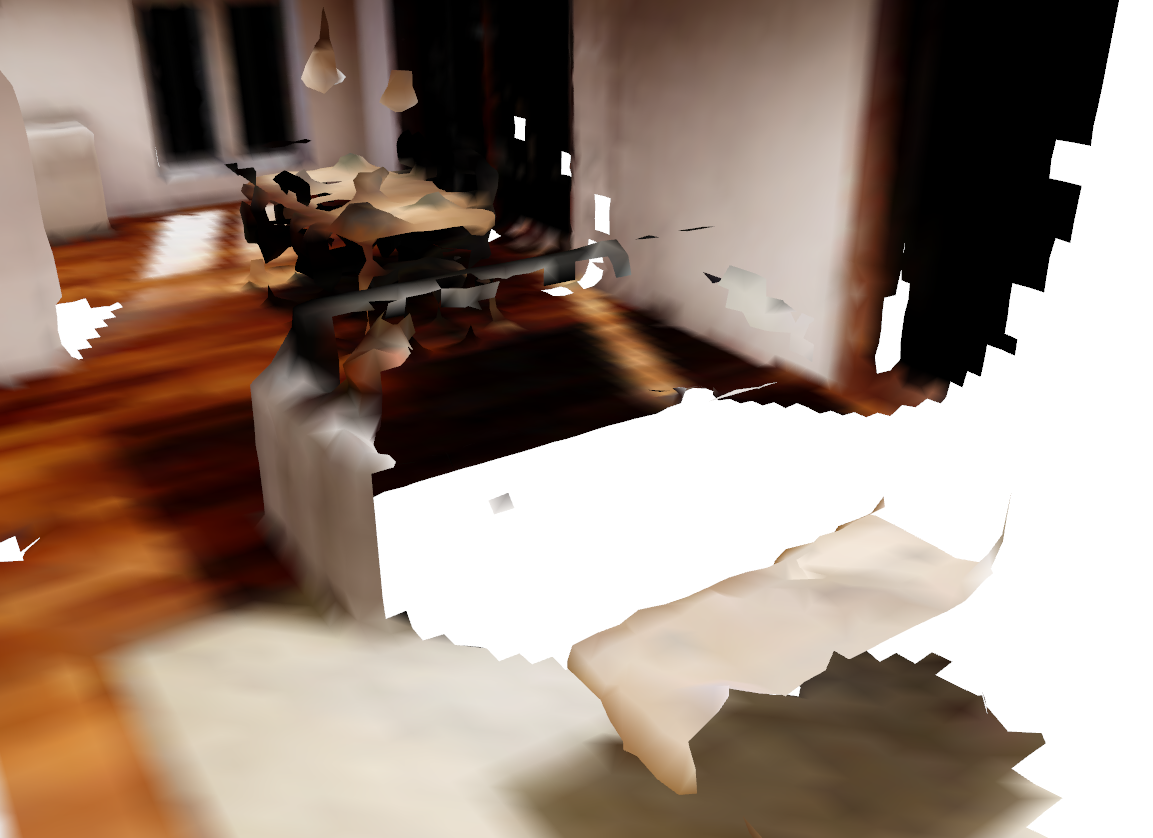}
\includegraphics[width=0.49\linewidth]{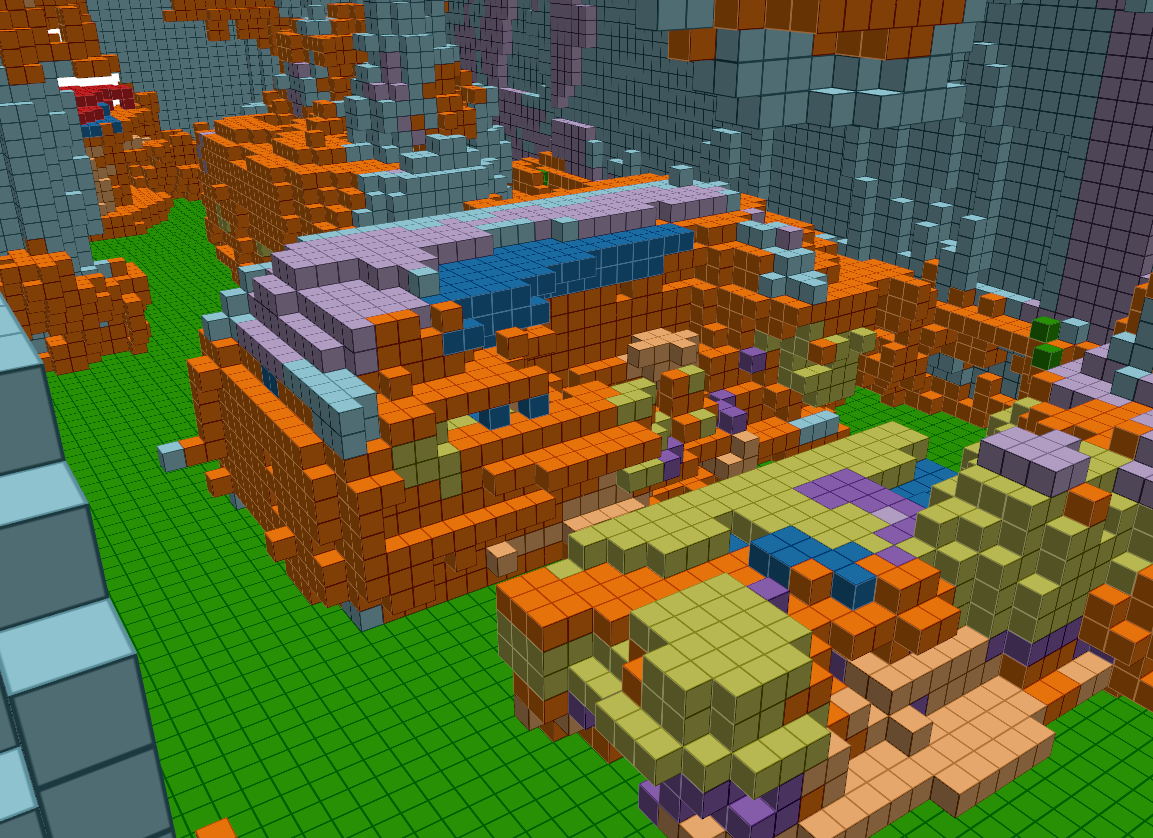} \\ \vspace{3pt} 
\includegraphics[width=0.49\linewidth]{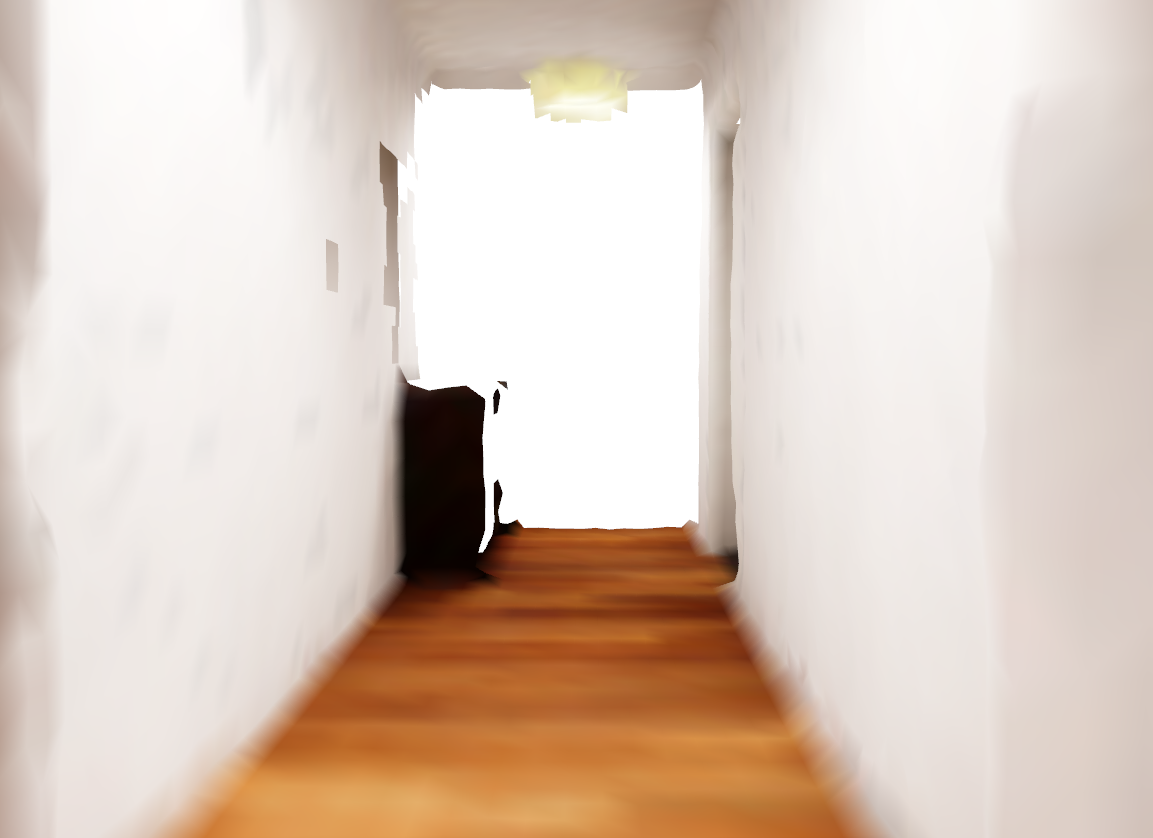}
\includegraphics[width=0.49\linewidth]{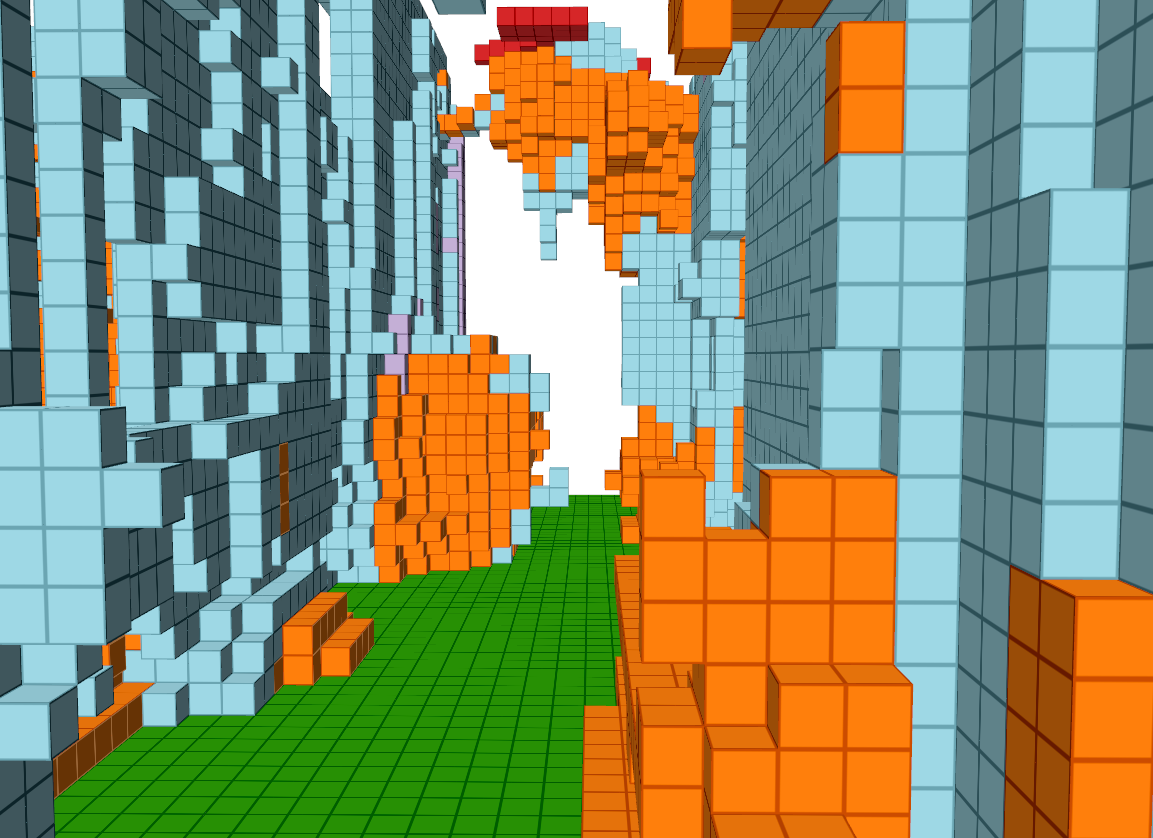}

\caption{Qualitative scene completed mapping results, showing the TSDF-based measured mesh (left) and voxel-wise SC layer (right). Top: Successful SC completing the occluded couch and table as well as the walls and floor. Bottom: Noisy SC predicting some clutter and potential occlusions in a hallway. }
\label{fig:qualitative}
\end{figure}

\setlength{\tabcolsep}{2pt}
\begin{table}
    \centering
    \caption{Performance metrics when planning on different maps.}
    \begin{adjustbox}{width=\linewidth}
    \begin{tabular}{lcccccc}
    Method & $\mathcal{T}_{\mathcal{C}=80}$ & $\mathcal{T}_{\mathcal{C}=50}$ & $\mathcal{T}_{\mathcal{M}=80}$ & $\mathcal{T}_{\mathcal{M}=50}$ & $E_\mathcal{C}$ & $E_\mathcal{M}$ \\
    \midrule
Optimistic (Ours) & \textbf{9.5} & 3.7 & \textbf{10.0} & \textbf{6.8} & \textbf{65.5} & \textbf{60.8} \\
Confidence (Ours) & 10.1 & 3.6 & 10.4 & 7.0 & 63.9 & 58.7 \\
SC-Blocking (Ours) & 10.9 & \textbf{3.2} & 10.8 & 6.9 & 64.4 & 59.0 \\
SC Layer & -$^\ast$ & 3.4 & -$^\ast$ & 9.2$^\ast$ & 55.7 & 44.8 \\
SCFusion & -$^\ast$ & 8.6 & 13.2$^\ast$ & 8.2$^\ast$ & 49.6 & 53.2 \\
Exploration & -$^\ast$ & 8.2 & 10.8$^\ast$ & 7.6 & 50.3 & 56.4 \\
    \midrule
    \multicolumn{7}{c}{\makecell{\footnotesize 
    Average time $\mathcal{T}_{(\cdot)=Y}$ in [min] till coverage $\mathcal{C}$ or measured volume $\mathcal{M}$ \\
    of $Y$ in [\%] is reached and expected $E_{(\cdot)}$ $\mathcal{C}$ and $\mathcal{M}$ in [\%]. Best \\
    number shown in bold. $^\ast$ Indicates not all runs reached the goal metric. }}
    \end{tabular}
    \end{adjustbox}
    \label{tab:map_exp}
\end{table}

A major role of the mapping system is to enable efficient planning.
We evaluate the capacity of different mapping approaches to facilitate exploration planning by running an identical exploration planner using the classical volumetric gain of Eq.~(\ref{eq:gain_exploration}) on different mapping approaches.
We evaluate three configurations considering a feature of our proposed mapping system (Sec.~\ref{sec:approach_sc_planning}) each.
\emph{Optimistic} allows traversal of predicted free space but uses classical ray-casting.
\emph{Confidence} is identical to Optimistic, but with a moderate confidence of threshold $\tau_c=\num{10}$\%.
\emph{SC-Blocking} uses conservative collision checking, but blocks rays if voxels are predicted occupied.
We compare these methods against planning on \emph{SCFusion} \cite{wu_scfusion_2020}, voxblox \cite{Oleynikova2017VoxbloxI3}, and the \emph{SC Layer}, as in Sec.~\ref{sec:results_map_type}.
Note that a classical planner using only the measured map of voxblox results in a traditional \emph{Exploration} pipeline as presented in \cite{Schmid20ActivePlanning}.

The progress in coverage $\mathcal{C}$ of planners using different maps is shown in Fig.~\ref{fig:map_exp} (left), highlighting the notable impact of the mapping approach.
First, we observe that planners using our multi-layer map achieve the highest performance, as they converge more quickly to the final $\mathcal{C}$ than Exploration, while also resulting in a higher final $\mathcal{C}$.
Confidence tends to be marginally slower than Optimistic, as less information is available for planning.
SC-Blocking tends to be a bit slower, as few erroneous occupied voxels can block an entire view, potentially outweighing the benefits of more accurate observable volume estimation when using the employed network, shown in Fig.~\ref{fig:qualitative}.
This effect is significantly stronger for SCFusion and SC Layer, which even though they add additional SC data to their map achieve lower performance than Exploration.
This shows that added data through SC can also have adverse effects and thus requires careful consideration.

To investigate the impact of the map on planning separately from the added value of using SC, the volume that was measured by the sensor $\mathcal{M}$ is depicted in Fig.~\ref{fig:map_exp} (right).
Here the same effect can be observed.
While the added occupancy data in SCFusion guarantees safety of the robot, it can hinder accurate gain estimation, resulting in sub-optimal paths that measure less of the environment.
The effect is even stronger for SC Layer, where incorrect occupied voxels can further hinder navigation.
Due to the hierarchical map representation and ability to refine occupied predictions, our methods are less affected and measure at least as much as Exploration.

The planner performance both on $\mathcal{C}$ and $\mathcal{M}$ is further analyzed in Tab.~\ref{tab:map_exp}, showing the average time $\mathcal{T}_{(\cdot)}$ it takes a planner to reach an exploration goal as well as the expected performance $E_{(\cdot)}$.
While our approaches are able to quickly cover the scene, interestingly, we find that running a classical exploration planner on our SC-aware maps tends to lead to paths that also measure more of the environment more quickly.
This is reflected in the high values of $\mathcal{T}_\mathcal{M}$ and $E_\mathcal{M}$.

While we find that optimistic planning in Optimistic and Confidence can be beneficial for quicker exploration, it is important to point out that one run of Confidence encountered a collision after $\SI{10}{min}$.
Although Confidence is by construction at least as safe as Optimistic and only one collision occurred, this highlights that safety can no longer be guaranteed, further discussed in Sec.~\ref{sec:discussion}.


\subsection{Impact of Information Gains}

\begin{figure}
\centering
\includegraphics[width=\linewidth]{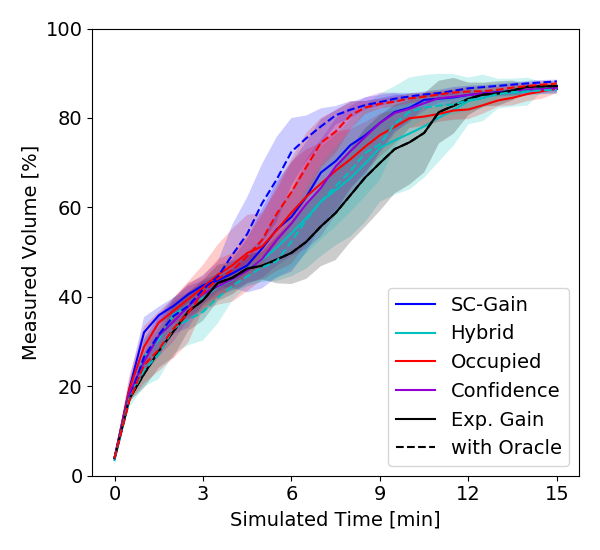}
\caption{Measured volume $\mathcal{M}$ over time for different information gain formulations (color). Solid lines indicate the performance using network predictions, whereas dashed lines show the performance if the scene completions were perfect.}
\label{fig:ig}
\end{figure}

\begin{table}
    \centering
    \caption{Performance of different information gains using our network (left) and ground truth (right) as SC predictions.}
    \begin{adjustbox}{width=\linewidth}
    \begin{tabular}{lcccccc}
    \multirow{2}{*}{Method} & \multicolumn{3}{c}{Network} & \multicolumn{3}{c}{Oracle} \\
    & $\mathcal{T}_{\mathcal{M}=80}$ & $\mathcal{T}_{\mathcal{M}=50}$ & $E_\mathcal{M}$ & $\mathcal{T}_{\mathcal{M}=80}$ & $\mathcal{T}_{\mathcal{M}=50}$ & $E_\mathcal{M}$ \\
    \midrule
SC-Gain & \textbf{9.0} & 6.0 & \textbf{63.8} & \textbf{7.6} & \textbf{4.8} & \textbf{67.2} \\
Hybrid & 10.3 & 6.9 & 60.5 & 10.1 & 6.4 & 60.3 \\
Occupied & 11.1 & \textbf{5.4} & 62.3 & 8.2 & 5.2 & 65.0 \\
Confidence & 9.2 & 5.9 & 63.2 & \textbf{7.6} & \textbf{4.8} & \textbf{67.2} \\
Exp. Gain & 10.8 & 6.9 & 59.0 & 10.8 & 6.9 & 59.0 \\
    \midrule
    \multicolumn{7}{c}{\makecell{\footnotesize 
    Average time $\mathcal{T}_{\mathcal{M}=Y}$ in [min] till volume $\mathcal{M}$
    of $Y$ in [\%] is \\measured and expected $E_\mathcal{M}$ in [\%]. Best number shown in bold.}}
    \end{tabular}
    \end{adjustbox}
    \label{tab:ig}
\end{table}

In order to evaluate the effectiveness of the information gains proposed in Sec.~\ref{sec:approach_gains}, we deploy an identical planner where only the gain is varied.
All methods operate on our default hierarchical map, planning conservatively, being non-blocking, and using $\tau_c=0$\%.
To assess the practical performance as well as the potential of each gain formulation, each method is run on our map fusing predictions of the network, and on a map using the ground truth provided by an oracle as the SC layer.
Although the second case can not be deployed on a real robot, it highlights the potential performance of each gain if the scene completions were perfect.

We show the measured volume $\mathcal{M}$ of each planner over time in Fig.~\ref{fig:ig}, with the corresponding performance metrics being summarized in Tab.~\ref{tab:ig}. 
Notably, there is a clear separation between the SC aware methods \emph{SC-Gain}, \emph{Occupied}, and \emph{Confidence} and the traditional volumetric \emph{Exploration} Gain.
\emph{Hybrid} places in between these methods and Exploration.
This further highlights the capability of SC for efficient planning, as the SC-only gains outperform the gains including exploration components.
This can be explained, as the SC act as a region of interest predictor. 
For example, the scene completions typically do not extend far beyond walls, which makes peeking through a small hole in the wall less desirable.
Lastly, Occupied shows reduced performance after 10 min, as few remaining predicted occupied voxels can compromise planning performance.

Interestingly, when using ground truth predictions, the performance of hybrid stays almost the same.
On the other, the pure SC-methods show significantly improved performance. 
Note that Confidence and SC-Gain become identical when the oracle is used, as the confidence is always 100\%.
The stark contrast between exploration and SC-Gain with oracle demonstrates the capability of SC predictions to positively influence planning, speeding up complete measurement of the scene $\mathcal{T}_{\mathcal{M}=80}$ with the sensor by up to 30\%.
Furthermore, even though the SC network is not the focus of this work and achieves only moderate performance (Tab.~\ref{tab:network}), this is already sufficient to improve planning performance.
In particular, SC-Gain, Occupied, and Confidence realize over 50\% of their improvement potential compared to Exploration when employing the presented network.
This is most pronounced for the proposed SC-Gain, speeding up exploration by 17\% compared to classical planning even when using the imperfect network.


\subsection{Complete SC-Aware Mapping and Planning System}

\begin{figure*}
\centering
\includegraphics[width=0.48\linewidth]{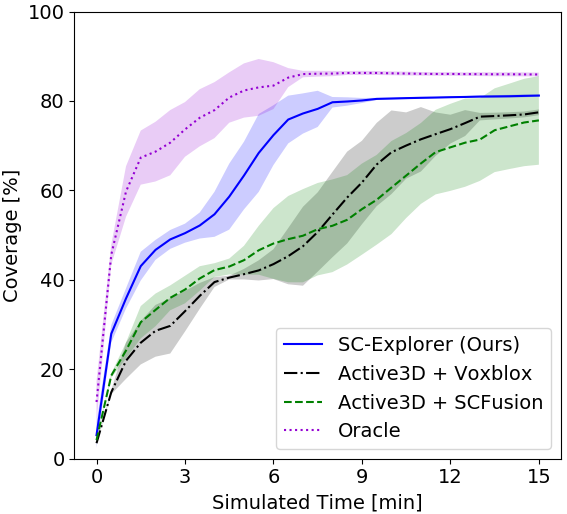}
\includegraphics[width=0.48\linewidth]{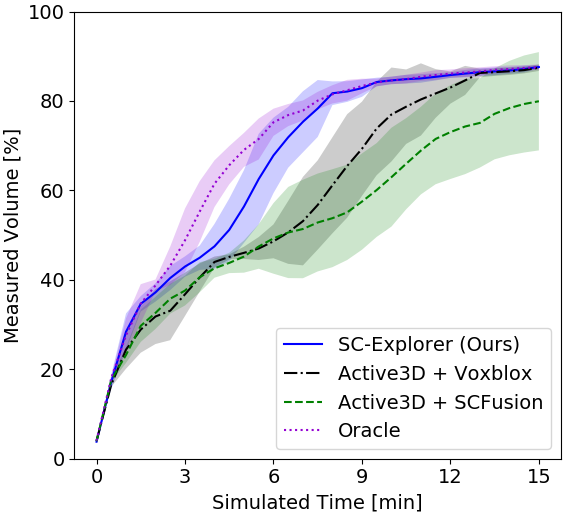}
\caption{Coverage $\mathcal{C}$ (left) and measured volume $\mathcal{M}$ (right) of SC-Explorer, our complete SC-aware mapping and planning system, compared to state of the art systems of Active3D \cite{Schmid20ActivePlanning} using voxblox \cite{Oleynikova2017VoxbloxI3} and SCFusion \cite{wu_scfusion_2020}, with Oracle as an upper performance bound.}
\label{fig:full}
\end{figure*}

\begin{table}
    \centering
    \caption{Performance of different exploration systems.}
    \begin{adjustbox}{width=\linewidth}
    \begin{tabular}{lcccccc}
    Method & $\mathcal{T}_{\mathcal{C}=80}$ & $\mathcal{T}_{\mathcal{C}=50}$ & $\mathcal{T}_{\mathcal{M}=80}$ & $\mathcal{T}_{\mathcal{M}=50}$ & $E_\mathcal{C}$ & $E_\mathcal{M}$ \\
    \midrule
SC-Explorer (Ours) & \textbf{8.5} & \textbf{3.4} & \textbf{8.0} & \textbf{5.1} & \textbf{67.5} & \textbf{66.4} \\
Active3D + Voxblox & -$^\ast$  & 8.2 & 10.8 & 7.6 & 50.3 & 56.4 \\
Active3D + SCFusion & 14.6$^\ast$ & 8.4$^\ast$ & 13.2$^\ast$ & 8.2$^\ast$ & 51.6 & 53.2 \\
Oracle & \underline{4.7} & \underline{1.2} & \underline{8.2} & \underline{3.6} & \underline{79.4} & \underline{69.1} \\
    \midrule
    \multicolumn{7}{c}{\makecell{\footnotesize $\mathcal{T}_{(\cdot)}$ in [min], $E_{(\cdot)}$ in [\%]. $^\ast$Indicates not all runs reached the target.\\Best number shown underlined (Oracle), second best shown in bold.}}
    \end{tabular}
    \end{adjustbox}
    \label{tab:full}
\end{table}

Eventually, we evaluate the overall performance of \emph{SC-Explorer}, our complete SC-aware mapping and planning pipeline. 
In particular, SC-Explorer uses the proposed mapping approach with non-blocking ray-casting and our SC-aware information gain $I_{sc}(v)$.
For maximal performance, we allow optimistic planning, and employ a moderate confidence threshold of $\tau_c=10$\% for safety and to focus on high confidence areas.
We compare against the \emph{Active3D} exploration system of \cite{Schmid20ActivePlanning}, using voxblox \cite{Oleynikova2017VoxbloxI3} as map and a classical volumetric exploration gain.
As a SC-aware baseline, we run Active3D with \emph{SCFusion} \cite{wu_scfusion_2020} as mapping system.
It is worth pointing out that, for fair comparison, all these approaches are based on an identical planner and SC-network.
Nonetheless, as our developed method can readily be employed also as a local planner with additional global planners as in \cite{schmid2021glocal, Selin_nbv_fron, history_nbvp}, with different sampling schemes such as \cite{dai2020fast, kompis2021informed, meng_2stage_expl}, or various SC networks \cite{palnet, Wang2019ForkNetMV, Zhang2019CascadedCP}, we expect similar performance benefits when our approach is extended and compared to other scenarios. 
As a final comparison, we include an \emph{Oracle} planner, that has from the start access to the complete ground truth map.
As information gain, it computes the true number of observable unmeasured voxels and is optimistic, i.e. can optimize its path globally \cite{Schmid20ActivePlanning}.
As the coverage of the oracle planner would trivially be $\num{100}$\%, it is evaluated as if it was equipped with our network at $\mathcal{P}=\num{100}$\%.
Nevertheless, it has access to the complete map for planning, and thus serves as an upper performance bound.

We show the coverage $\mathcal{C}$ and measured volume $\mathcal{M}$ of each system in Fig.~\ref{fig:full}, and summarize the performance metrics in Tab.~\ref{tab:full}.
First, we observe that SC-Explorer is able to cover more of the scene significantly quicker than both baselines.
Our system shows reliable and consistent performance, indicated by the comparably low standard deviation.
Notably, SC-Explorer converges to its final $\mathcal{C}$ extremely quickly, demonstrating a 73\% speed up over Active3D and taking only 15\% longer than the Oracle.

The difference in final $\mathcal{C}$ is explained by the added SC. 
Although eventually only little predicted volume is left unmeasured (c.f. Fig.~\ref{fig:fusion_coverage}), this can add additional information compared to Active3D.
However, the quality of these left over predictions could still be improved as indicated by the Oracle.
Similarly, adding completions to the map is not always necessarily beneficial, as planning progress is slightly slower when using SCFusion compared to voxblox.
Lastly, as the Oracle plans a globally optimized path, it exhibits a much smoother exploration curve without the plateaus resulting from having to traverse the already explored environment.

We find similar results when analyzing only the measured areas.
Due to its ability to predict the environment, and then plan optimistically to observe as much of these predictions, SC-Explorer is significantly closer to the Oracle performance.
While it does not quite manage to find the optimal path for shorter time budgets, its complete exploration time $\mathcal{T}_{\mathcal{M}=80}$ is on par with the oracle, showing a speed up of 35\% over Active3D + Voxblox and 65\% over Active3D + SCFusion.
These findings demonstrate the strong capability and importance of scene completions to predict the unknown areas around the robot during exploration planning. 
Even without considering the completions in the final map, they can be used to guide the robot to significantly quicker exploration.


\subsection{Ablation Study}

\begin{table}
    \centering
    \caption{Impact of different components on SC-Explorer.}
    \begin{adjustbox}{width=\linewidth}
    \begin{tabular}{lcccccc}
    Component & $\mathcal{T}_{\mathcal{C}=80}$ & $\Delta\mathcal{T}_{\mathcal{C}=80}$ &  $\mathcal{T}_{\mathcal{M}=80}$ & $\Delta\mathcal{T}_{\mathcal{M}=80}$ \\
    \midrule
    SC-Explorer & 8.5 & - & 8 & - \\
    w/o Information Gain & 10.7 & -20.6 & 10.4 & -23.1 \\
    w/o Optimistic Planning & 9.1 & -6.6 & 8.6 & -7.0 \\
    w/o Confidence & 8.7 & -2.3 & 8.5 & -5.9 \\
    w/ SC-blocking & 9.9 & -14.1 & 9.8 & -18.4 \\
    w/ Perfect Network & 5.7 & 49.1 & 7.6 & 5.3 \\
    \midrule
    \multicolumn{5}{c}{\makecell{\footnotesize $\mathcal{T}_{(\cdot)}$ in [min]. $\Delta(\cdot)$ is the improvement over SC-Explorer in [\%]. }}
    \end{tabular}
    \end{adjustbox}
    \label{tab:ablation}
\end{table}

\begin{figure*}
\centering
\vspace*{\fill}
\includegraphics[width=0.32\linewidth]{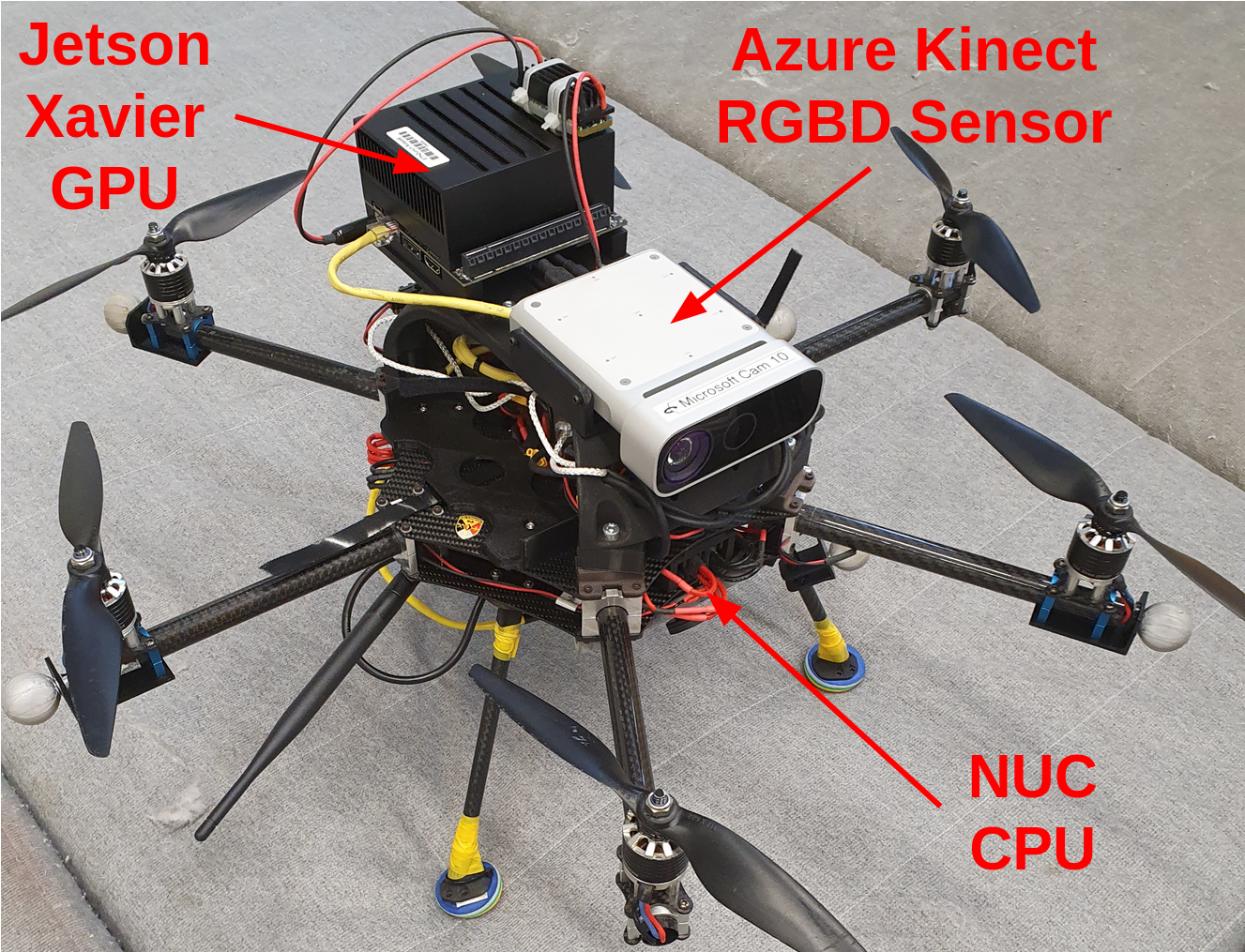}
\vspace*{\fill} 
\includegraphics[width=0.33\linewidth]{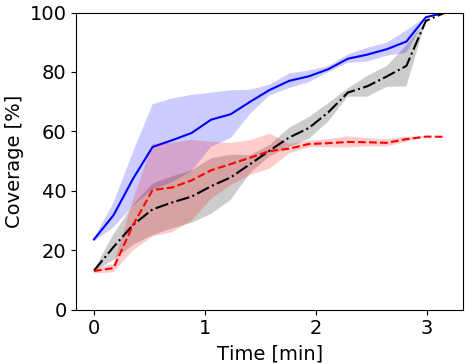}
\includegraphics[width=0.33\linewidth]{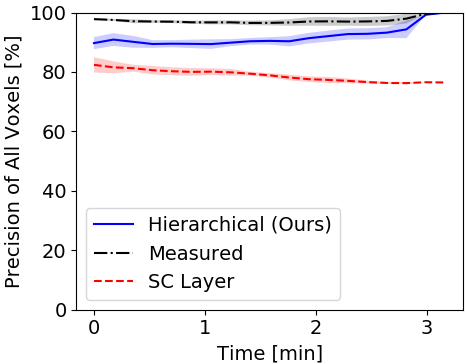}
\caption{Left: MAV equipped with mobile sensor, CPU, and GPU used for the robot experiments. Right: Quantitative evaluations of SC-Explorer. Similar to the simulation experiments, we find that even on the real robot the added scene completions can significantly improve the map coverage with only minimal reduction in map accuracy. Lastly, the added completions are sufficient to guide the robot to an efficient and complete exploration path. }
\label{fig:kiwi}
\end{figure*}

Lastly, we analyze the impact of all discussed components on the performance of SC-Explorer in an ablation study shown in Tab.~\ref{tab:ablation}. 
First, we observe that all mechanisms enabled by our multi-layer mapping and SC-aware planning system contribute to the performance of SC-Explorer.
In particular, the proposed information gain governing the robot behavior has the strongest impact on overall performance.
In a similar range is the effect of excluding predicted voxels from blocking ray-casting.
This is consistent with our previous findings, where the added occupied predictions in SCFusion can hinder planner performance.
While the effect of the other two components is less notable, the employed confidence layer lets the planner focus on high probability areas, and optimistically plan longer horizon paths in a larger action space around the robot, slightly boosting performance.
Lastly, having a perfect network would significantly improve the coverage induced by SC, but SC-Explorer is already within close range of the performance when measuring the environment.


\subsection{Robot Experiment}

We demonstrate the applicability of our method on a real MAV exploring an indoor scene.
Therefore, the hexacopter drone shown in Fig.~\ref{fig:kiwi} (left) is equipped with a Microsoft Azure Kinect RGBD sensor and the identical computational hardware used for the simulation experiments, described in Sec.~\ref{sec:hardware}.
We employ SC-Explorer to explore an U-shaped indoor room, as can be seen in the reconstruction in Fig.~\ref{fig:kiwi_qualitative}.
State estimates are provided by an external Vicon system for safety, although also on-board estimates can be used.
Otherwise the robot operates completely autonomously using only on-board sensing and computation.

We show qualitative results of SC-Explorer operating on the physical robot in Fig.~\ref{fig:kiwi_qualitative}.
We observe that our system is able to reconstruct the environment, and meaningfully completes holes in the reconstruction such in the floor below the robot starting position or in the top right corner. We further note that, even in spite of the large domain shift between the training dataset and this cluttered scene and sensor, SC-Explorer is able to meaningfully predict the yet unknown areas and guides the robot to plan and execute effective paths to measure these areas. 

Lastly, we present quantitative evaluations in Fig.~\ref{fig:kiwi} (right). 
Since accurate Vicon state estimates are employed, we use the most complete reconstruction at the end of the longest exploration run as approximate ground truth of the room.
Note that the planner was manually terminated around $\SI{3}{min}$ when exploration seemed complete.
We show the mean and standard deviation over 5 experiments.
As in the simulation experiments, we observe that the SC is able to notably speed up coverage of the environment, adding valuable information with only minimal decrease in map accuracy.
Also in line with the simulated results, we see the precision $\mathcal{P}$ improve over time as the robot measures more of the environment.

These experiments highlight the capability of our system, \emph{SC-Explorer}, to operate reliably on-board of a mobile MAV, using only on-board resources.
In addition, in spite of the notably different environment, sufficient SC are achieved to enable plans for rapid measurement and high quality coverage of the scene.


\section{Discussion}
\label{sec:discussion}

\minisec{Scene Completions in Mapping} 
As shown in our experiments, adding scene completions in the map can contribute significant amounts of true information, speeding up coverage by up to 73\% while dropping only 5\% in mapping accuracy. 
Importantly, as our multi-layer map explicitly represents which parts of the map were measured and which were predicted, the completed areas can later on also be verified or corrected by a human operator or post-processing operations.
Notably, small holes in the map are typically easy for humans to fill in. 
If these are completed, the robot is incentivized to move on rather than measure every last bit, and thus explores the larger scale structure of a scene more quickly. 
Such behavior can be desirable for rapid initial screening instead of thorough inspection, such as in a disaster response scenario.

In either case, adding SC to the map has significant impact on robot planning performance.
First, it can be highly beneficial to predict the areas to inspect, speeding up measurement of an environment with the sensor by 35\%.
However, it can also block the predicted views or robot navigation, leading to sub-optimal paths or the robot even getting stuck, hindering exploration progress compared to SC-unaware methods.
How SC is integrated into the map and used for planning thus requires careful consideration.

\minisec{Optimistic Planning and Safety} 
Allowing the robot to plan ahead into predicted free space can further speed up performance, up to 7\% in our experiments.
Although optimistic planning is potentially unsafe, we did not observe notably more collisions. 
Out of the 80 optimistic experiments conducted, only 3 collided (3.8\%), whereas for the conservative methods 5 out of 110 collided (4.6\%).
This demonstrates that also a classical TSDF map is not perfectly safe as, due to discretization artifacts, only 90\% precision is achieved.
Furthermore, TSDFs are not able to represent thin structures well, which are frequently encountered in complex indoor environments.
However, this issue can potentially be alleviated by employing approaches such as multi-resolution maps \cite{schmid_panoptic_2022, vespa2019adaptive} or infinite resolution deep implicit maps with shape priors \cite{lionar_neuralblox_2021,huang2021di}.
Lastly, navigation in narrow and cluttered indoor environments is a challenging problem, where also imprecision of the controller and other parts of the pipeline that are not related to mapping and planning can lead to collisions.

While the incidence rate of collisions is too low for a meaningful experimental study, we note that the collision rate is comparable to the conservative approach, but also observe that there is a comparably small performance gain when planning optimistically.
This can be explained by several reasons.
First, since the robot plans to measure the environment with its sensor, the measured map changes at every time step, thus making the planning problem inherently short horizon.
Similarly, even if the robot plans far ahead, the executed path segment before re-planning is always well within the sensing range.
As our hierarchical map prioritizes the measured over the predicted map, as long as the robot faces the direction of travel it typically stays within the measured area, potentially explaining the low collision rate.
Furthermore, while the precision of the predicted map is only 50-75\%, the voxels are highly correlated. 
Thus, an object that is slightly inflated or deflated in size is less problematic than objects that are entirely missed or hallucinated.
While having different confidence thresholds can potentially improve safety, we note that the failure cases of classical maps are generally better understood than those of neural networks, which might make safety management easier when conservative planning is employed.

\minisec{Blocking Ray-casting} 
Using the predicted occupied areas to block rays during information gain computation has the potential to more accurately estimate the areas that will be observed.
If the SC is highly accurate as in the oracle case, this can lead to further speedup of exploration before convergence $\mathcal{T}_{\mathcal{M}=70}$ by up to 17\%.
However, for the network employed in this work, the performance of SC-Explorer drops by 18\% in $\mathcal{T}_{\mathcal{M}=80}$ when using SC-blocking.
In summary, whether or not SC-blocking improves performance mostly depends on the expected network accuracy and characteristics.
However, since noisy predictions have the potential to completely block the robot and compromise performance, non-blocking ray-casting is likely the more robust option for most applications. 
This highlights a strength of the proposed multi-layer mapping approach, as it enables detailed configurations on how SC should be modeled and used in planning. 
Alternatively, parts of the potential of blocking ray-casting is already realized by our information gain, as the network predicts areas that are interesting to observe and thus influences planning in a similar way.

\minisec{Exploration Planning using SC}
Even if an application requires complete measurement of a scene or chooses not to rely on SC in the final map, the extension of the environment using deep learnt shape priors adds valuable information for planning.
In particular in exploration planning, the majority of the relevant information to select the NBV is unknown.
While most current methods rely on heuristics such as that all unknown space will be observable, SC can provide valuable expectations of the unknown for planning.
This is reflected in the stark performance increase of up to 35\% speedup in measuring the environment compared to an SC-unaware planner.
Furthermore, the presented incorporation of SC results in a robust and interpretable way to combine the power of deep learning methods with exploration planning.
The added SC implicitly adapt the planning behavior to the current environment, such as following a straight hallway, as long as the network operates reliably.
Lastly, the presented approach allows trading-off all the discussed factors, such that the system can be guaranteed to be as safe as the measured map, and at least as performing as exploration when using the hybrid gain, although potentially sacrificing performance for reliability. 


\section{Conclusions}

In this work, we have explored the use of incremental 3D scene completion for exploration mapping and planning. 
We investigated different fusion strategies, showing that completions require different fusion approaches as objects can easily be missed. 
We further find that adding SC to the robot map can significantly speed up accurate coverage of an environment, but can also be hindering planning performance.
We thus presented SC-Explorer, a complete SC-aware exploration mapping and planning pipeline.
Our approach incrementally fuses scene completions into a hierarchical multi-layer map, guaranteeing safety of the robot and facilitating efficiency of the planner.
Lastly, a novel planning system and information gain allows trading off safety vs. efficiency, and combined with an SC-aware information gain enables quick exploration planning.
We showed in thorough experimental evaluations, that SC-Explorer can speed up coverage by 73\% compared to classical exploration systems, or increase the rate at which the environment is measured with the robots sensors by 35\%.
We validate the our method on a fully autonomous MAV, using only on-board sensing and computation to complete the scene and plan efficient paths, showing reliable and rapid exploration even in cluttered and complex scenes.

We believe this work further gives interesting directions for future research.
First, we showed that innovation in 3D scene completion and more accurate predictions can still improve planning performance.
In addition, while the presented occupancy fusion is a first promising direction, there is notable potential for better uncertainty estimates of scene completions and their modeling in the map.
While we employ a semantic loss to train more accurate shape completions of the network and make use of them to calibrate the fusion uncertainty, extensive fusion of semantics in the map and their use for planning could further improve performance and interpretability.
This also leaves space for further innovation on information gains, for example also accounting for the semantics or which parts of the map will be completed.
Lastly, we believe that the presented framework can bring the predictive power of deep neural networks into various informative path planning tasks other than exploration.


\section*{ACKNOWLEDGMENT}
We are grateful for Mike Allenspach's support with the robot experiments, and fruitful discussions with Marius Fehr.


{\small
\bibliographystyle{IEEEtran}
\bibliography{IEEEfull,references}
}


\end{document}